\documentclass[10pt,twocolumn,letterpaper]{article}

\usepackage{iccv}
\usepackage{times}
\usepackage{amsmath}
\usepackage{amssymb}
\usepackage{booktabs}
\usepackage{graphicx}
\usepackage[labelfont=bf,format=plain,labelformat=simple,labelsep=period,font=small,compatibility=false]{caption}
\usepackage[font=footnotesize,skip=0pt,subrefformat=parens]{subcaption}

\usepackage{amsmath,amsfonts,bm}

\def\eqref#1{equation~\ref{#1}}

\def\1{\bm{1}}

\def\rvg{{\mathbf{g}}}

\def\rvy{{\mathbf{y}}}

\def\rmA{{\mathbf{A}}}

\DeclareMathAlphabet{\mathsfit}{\encodingdefault}{\sfdefault}{m}{sl}
\SetMathAlphabet{\mathsfit}{bold}{\encodingdefault}{\sfdefault}{bx}{n}

\def\sR{{\mathbb{R}}}

\DeclareMathOperator*{\argmin}{arg\,min}

\usepackage[pagebackref=true,breaklinks=true,letterpaper=true,colorlinks,bookmarks=false]{hyperref}

\usepackage[capitalize]{cleveref}
\crefname{section}{Sec.}{Secs.}
\Crefname{section}{Section}{Sections}
\Crefname{table}{Table}{Tables}
\crefname{table}{Tab.}{Tabs.}

\usepackage{url}
\usepackage{amsthm}
\usepackage{thmtools}
\usepackage{mathtools}
\usepackage{multirow}
\usepackage[accsupp]{axessibility}

\graphicspath{ {images/} }

\iccvfinalcopy 

\def\iccvPaperID{1893} 

\ificcvfinal\fi

\makeatletter
\def\thm@space@setup{\thm@preskip=0pt
\thm@postskip=0pt}
\makeatother

\newtheorem{theorem}{Theorem}

\newtheorem{lemma}[theorem]{Lemma}
\theoremstyle{definition}
\newtheorem{definition}{Definition}
\newtheorem{remark}{Remark}[theorem]

\declaretheoremstyle[%
  headfont=\normalfont\itshape,%
  postheadspace=0.5em,%
  qed=\qedsymbol%
]{mystyle} 
\declaretheorem[name={Proof},style=mystyle,unnumbered,
]{prf}

\DeclareMathOperator*{\Sim}{Sim}
\DeclareMathOperator*{\Dist}{Dist}
\DeclareMathOperator*{\Reg}{Reg}

\newcommand{\normtwotwo}[1]{\vert \vert{#1}\vert \vert^2_2}
\newcommand{\normL}[2]{\vert \vert{#1}\vert \vert_{L^{#2}(H^\star)}}
\newcommand{\snormtwo}[1]{\vert {#1}\vert^2}
\newcommand{\st}{\text{s.t.}\quad}
\newcommand{\where}{\text{where}\quad}

\begin{document}

\title{Towards Saner Deep Image Registration}

\author{
\qquad Bin Duan\qquad\qquad\, Ming Zhong\quad Yan Yan\thanks{Corresponding author}\\
\tt\small{bduan2@hawk.iit.edu\qquad\{mzhong3,\,yyan34\}@iit.edu} \\
Illinois Tech, Chicago, IL 60616, USA\\
\small{\url{https://github.com/tuffr5/Saner-deep-registration}}
}

\maketitle
\thispagestyle{empty}

\begin{abstract}
With recent advances in computing hardware and surges of deep-learning architectures, learning-based deep image registration methods have surpassed their traditional counterparts, in terms of metric performance and inference time. However, these methods focus on improving performance measurements such as Dice, resulting in less attention given to model behaviors that are equally desirable for registrations, especially for medical imaging. This paper investigates these behaviors for popular learning-based deep registrations under a sanity-checking microscope. We find that most existing registrations suffer from low inverse consistency and nondiscrimination of identical pairs due to overly optimized image similarities. To rectify these behaviors, we propose a novel regularization-based sanity-enforcer method that imposes two sanity checks on the deep model to reduce its inverse consistency errors and increase its discriminative power simultaneously. Moreover, we derive a set of theoretical guarantees for our sanity-checked image registration method, with experimental results supporting our theoretical findings and their effectiveness in increasing the sanity of models without sacrificing any performance.
\end{abstract}
\section{Introduction}

Learning maps between images or spaces, i.e. registration, is an important task, and has been widely studied in various fields, such as computer vision~\cite{Fu_2021_CVPR, Qin_2022_CVPR}, medical imaging~\cite{hering2022learn2reg, zhou2021deep}, and brain mapping~\cite{qu2022cross, wang2021bi}. With recent advances in modern computing hardware and deep-learning techniques, learning-based deep image registration methods have surpassed their traditional counterparts, both in terms of metric performance and inference time. Different from the traditional style of optimizing on single image pair~\cite{christensen1997volumetric, pennec1999understanding, beg2005computing, christensen2006introduction, avants2008symmetric, modat2010fast, heinrich2015multi} using diffeomorphic formulations, such as elastic~\cite{bajcsy1989multiresolution, shen2002hammer}, fluid mechanics~\cite{beg2005computing, hart2009optimal, vercauteren2009diffeomorphic} or B-spline~\cite{rueckert1999nonrigid}, existing deep registrations~\cite{sokooti2017nonrigid, balakrishnan2019voxelmorph, dalca2019unsupervised, mok2021conditional, kim2021cyclemorph, chen2021vit, siebert2021fast, lv2022joint, chen2022transmorph} focus on maximizing image similarities between transformed moving images and fixed images. Despite the effectiveness of this approach, it inevitably leads to over-optimization of image similarities and thus introduces non-smooth mappings~\cite{balakrishnan2019voxelmorph, zhang2018inverse, Greer_2021_ICCV, chen2022transmorph}, where smooth transformation maps are typically desirable, especially in the medical imaging domain.

\begin{figure}[!t]
    \centering
    \hspace{-1.4em}
    \includegraphics[width=0.915\linewidth]{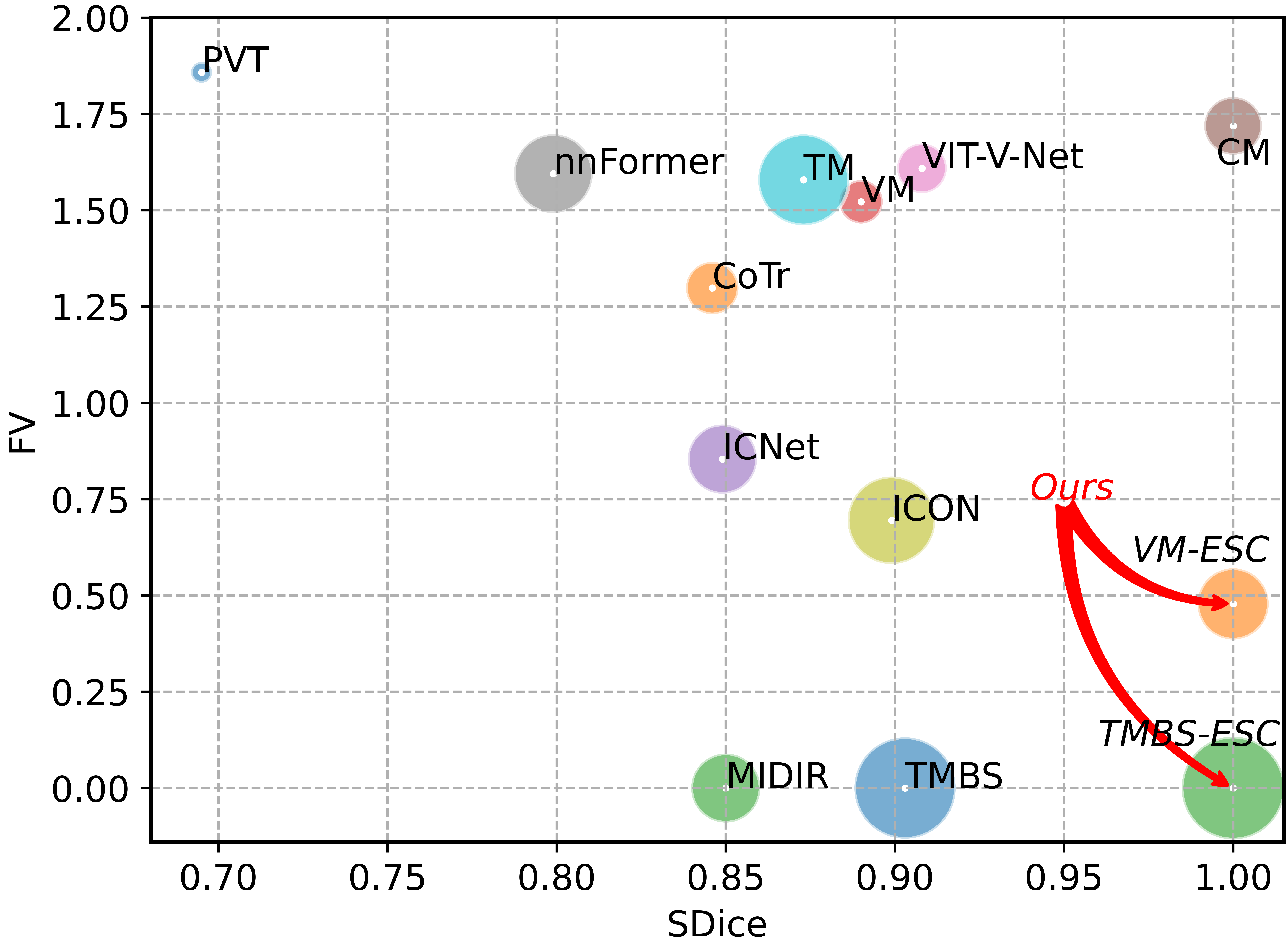}
    \caption{FV-SDice-Dice comparisons of deep registrations on IXI Brain dataset. The vertical axis is FV (\% of folded voxels), the horizontal axis is SDice (Self-Dice), and the circle size is Dice. Both sanity-checked models (\textit{VM-ESC} and \textit{TMBS-ESC}) achieve better diffeomorphism, competitive registration performance, and significantly improved self-sanity, compared to other models, including models with inverse consistency (ICNet~\cite{zhang2018inverse}, ICON~\cite{Greer_2021_ICCV}).}
    \label{fig:intro_checks}
\end{figure}
To tackle the over-optimized issue, popular remedies~\cite{bajcsy1989multiresolution, rueckert1999nonrigid, vialard2012diffeomorphic, schmah2013left, yang2017quicksilver, wang2020deepflash, shen2019networks, shen2019region} utilize add-ons such as Large Deformations Diffeomorphic Metric Mapping (LDDMM)~\cite{beg2005computing}, vector Stationary Velocity Field (vSVF)~\cite{shen2019networks}, B-spline~\cite{rueckert1999nonrigid}, Elastic~\cite{bajcsy1989multiresolution} or Demons~\cite{vercauteren2009diffeomorphic} to enforce diffeomorphism, requiring costly and iterative numerical optimizations~\cite{cao2018deformable, yang2017quicksilver, shen2019networks}. Other methods~\cite {dalca2019unsupervised, grzech2022variational} seek probabilistic formulation for the registration but can lead to inferior performance~\cite{chen2022transmorph}. Nonetheless, these methods operate only on one mapping in the direction from moving to fixed images, yet disregarding the relationship between different mappings from both directions, as shown in Appendix\footnote{Appendix goes \url{https://arxiv.org/pdf/2307.09696.pdf}}~\cref{fig:mapping}.

Recent studies~\cite{zhang2018inverse, Greer_2021_ICCV, mok2022unsupervised} have made great progress in modeling the relationship, i.e. inverse consistency, for different mappings. \cite{zhang2018inverse, Greer_2021_ICCV} explicitly enforce the relationship in a strict form for ideal inverse consistency. However, for applications such as brain tumor registration~\cite{baheti2021brain}, where there are regions with no valid correspondences, it is impractical to apply in such a strict manner. To select valid correspondences, \cite{mok2022unsupervised} utilize mean errors of similar anatomical locations as thresholds. However, it is always tricky to determine similar anatomical locations, especially for unsupervised registrations in medical imaging~\cite{balakrishnan2019voxelmorph, dalca2019unsupervised, siebert2021fast}. Different from previous methods~\cite{zhang2018inverse, Greer_2021_ICCV, mok2022unsupervised}, we introduce two straightforward yet effective constraints, namely, self-sanity check to reduce error while registering identical pairs and cross-sanity check to ensure inverse consistency. Using our sanity checks, we test a wide range of registration models and find that despite better performance such as Dice, most models are suffering from over-optimization of image similarities, leading to folded transformations and low sanity awareness, as shown in \cref{fig:intro_checks}. Moreover, our sanity checks can be applied to different registration models, where the extensive experiments certify that our method improves their sanity awareness, without sacrificing any performance.

Our findings are five-fold: (1) We find that despite better performance such as Dice, most models produce non-smooth transformations and are less aware of sanity errors due to over-optimization on image similarities; (2) We propose two novel sanity checks for the registration model training and derive corresponding theoretical foundations; (3) Our sanity checks not only help reduce the sanity errors of existing models but also assist them to produce more regular maps without diffeomorphic add-ons; (4) Our proposed sanity-checks are model-agnostic. It can be deployed to various models and is only needed during training so that it does generate no side effects for inference; (5) Experiments on IXI~\cite{IXI}, OASIS~\cite{marcus2007open}, and BraTSReg~\cite{baheti2021brain} datasets verify our findings and show on par or better performance.
\section{Background and Related Work}
\label{sec:rel}
\textbf{Background.} Consider a set of images defined on the domain $\Omega \subseteq \sR^n$ ($n$-D spatial domain) of some Hilbert space $H$. Let $f$ be the fixed, and $m$ be the moving $\in \Omega$. Let $u:\Omega\rightarrow \sR^n$ be a displacement vector field that maps from $H\times H \rightarrow H^\star$ (dual space of $H$) such that $\varphi = u + p$, where $p$ is the image grid of coordinates. $\varphi$ denotes the transformation from $m \rightarrow f$. The goal of registration is to compute such transformation $\varphi$, whose quality can be measured by some similarity functions in the form of $\Sim (f, m\circ\varphi)$. $m \circ \varphi$ denotes that $m$ is warped by $\varphi$. Without loss of generality, $\Sim(\cdot)$ can be replaced by any differentiable similarity function, such as Normalized Cross Correlation (NCC) and Mutual Information (MI). Alternatively, we can use other distance functions such as sum squared error (SSD) by replacing $-\Sim(\cdot)$ with $\Dist(\cdot)$. Hereafter, we will stick to the similarity notation for later derivations.

Let $g$ be the mapping function, parameterized using the model, where $u=g^{m\rightarrow f}\coloneqq g(m, f)$ stands for the displacement map from $m\rightarrow f$. In terms of such $g$ learned by the model, we assume that the similarity operator is concave, making $-\Sim$ convex when it is not a distance operator for a $(m, f)$ pair. Therefore, the optimization in standard learning-based registrations can be formulated as
\begin{equation}
    \min -\Sim(f, m \circ (g^{m\rightarrow f}+p)) + \lambda_r\normtwotwo{\Reg(g^{m\rightarrow f})},
    \label{eq:std_mini_subs}
\end{equation}
trying to find such $g$ for a $(m,f)$ pair in the mapping search space. Here, $\Reg(\cdot)$ term is added to smooth the transformation, penalizing sudden changes in the map. Most commonly, $\Reg(\cdot)$ can be in the form of $\nabla(u)$ (first-order spatial derivative) or $\nabla^2(u)$ (second-order spatial derivative), where $L^2$ norm of the image gradient is generally adopted in medical registration, resulting in a $H^1$ regularization. 
\begin{table}[!t]\small
    \centering
    \scalebox{1}{
    \begin{tabular}{|c|c|c|}
        \hline
        Method &  Relaxation/Approximation & Error Bound\\
        \hline
        ICNet~\cite{zhang2018inverse} & $||g^{m\rightarrow f}+\tilde{g}^{f\rightarrow m}||_{F}^2$ & -\\
        \hline
        \multirow{2}{*}{ICON~\cite{Greer_2021_ICCV}} &$||\varphi^{m\rightarrow f}\circ\varphi^{f\rightarrow m}-id||_2^2+$ &\multirow{2}{*}{-}\\
        &$
        \epsilon^2\normtwotwo{d \varphi^{m\rightarrow f}\sqrt{\rm Jac(\varphi^{m\rightarrow f})}}$&\\
        \hline
        \multirow{2}{*}{DIRAC~\cite{mok2022unsupervised}} & $|g^{m\rightarrow f}+\tilde{g}^{f\rightarrow m}|_2<$&\multirow{2}{*}{-}\\
        &$\frac{1}{N}\Sigma_N|g^{m\rightarrow f}+\tilde{g}^{f\rightarrow m}|_2+\beta$&\\
        \hline
        \multirow{2}{*}{Ours} & $|g^{m\rightarrow f}+\tilde{g}^{f\rightarrow m}|^2<$& \multirow{2}{*}{$\lambda_c(1-\alpha)\beta N$} \\
        &$\alpha(|g^{m\rightarrow f}|^2+|\tilde{g}^{f\rightarrow m}|^2)+\beta$&\\
        \hline
    \end{tabular}}
    \caption{Comparison between different inverse consistent registration methods. Due to the space limit, the formulations are shown only for $m\rightarrow f$, where $f\rightarrow m$ should follow the same way.}
    \label{tab:related}
\end{table}
\begin{definition}[Ideal/Strict inverse consistency] \label{def:inverse}
Given two different mappings: $g^{m \rightarrow f}$ and $g^{f \rightarrow m}$, if $\varphi^{m\rightarrow f}\circ\varphi^{f\rightarrow m}=id$, where $id$ denotes the identity transformation, we called these two mappings are strictly inverse consistent. The strict inverse consistency is equivalently formulated as $g^{m \rightarrow f} + \tilde{g}^{f \rightarrow m}=0$, where $\tilde{g}^{f \rightarrow m}$ is back-projected from $g^{f \rightarrow m}$.
\end{definition}

\textbf{Relation to other inverse consistent methods.} We show the relationship between methods in \cref{tab:related}. The formulation of ICNet~\cite{zhang2018inverse} follows strict inverse consistency in Def.~\ref{def:inverse} with Frobenius norm. Besides strict inverse consistency, ICON~\cite{Greer_2021_ICCV} adds regularization using the determinant of the Jacobian matrix of the transformation. Instead of explicitly enforcing strict consistency as in~\cite{zhang2018inverse, Greer_2021_ICCV}, DIRAC~\cite{mok2022unsupervised} and our approach both modulate the inverse problem with more relaxation. This inequality formulation allows us to tell whether a voxel has a valid correspondence, which is practically useful in registering image pairs with topological changes, e.g., brain tumor registration~\cite{mok2022unsupervised, mok2022robust}. Different from~\cite{mok2022unsupervised}, where means of inverse errors are utilized, we closely associate our formulation with the data, allowing us to show that the errors are upper bounded without resorting to extra information to determine similar anatomical structures, which has not been studied in other related works. Experimentally, our relaxation shows better performance over various metrics. Besides, to the best of our knowledge, our work is the first to consider the self-sanity error directly on displacements rather than for image similarities as in~\cite{kim2021cyclemorph} for medical image registration studies.
\section{Methodology}

\begin{figure}[!t]
    \centering
    \includegraphics[width=0.6\linewidth]{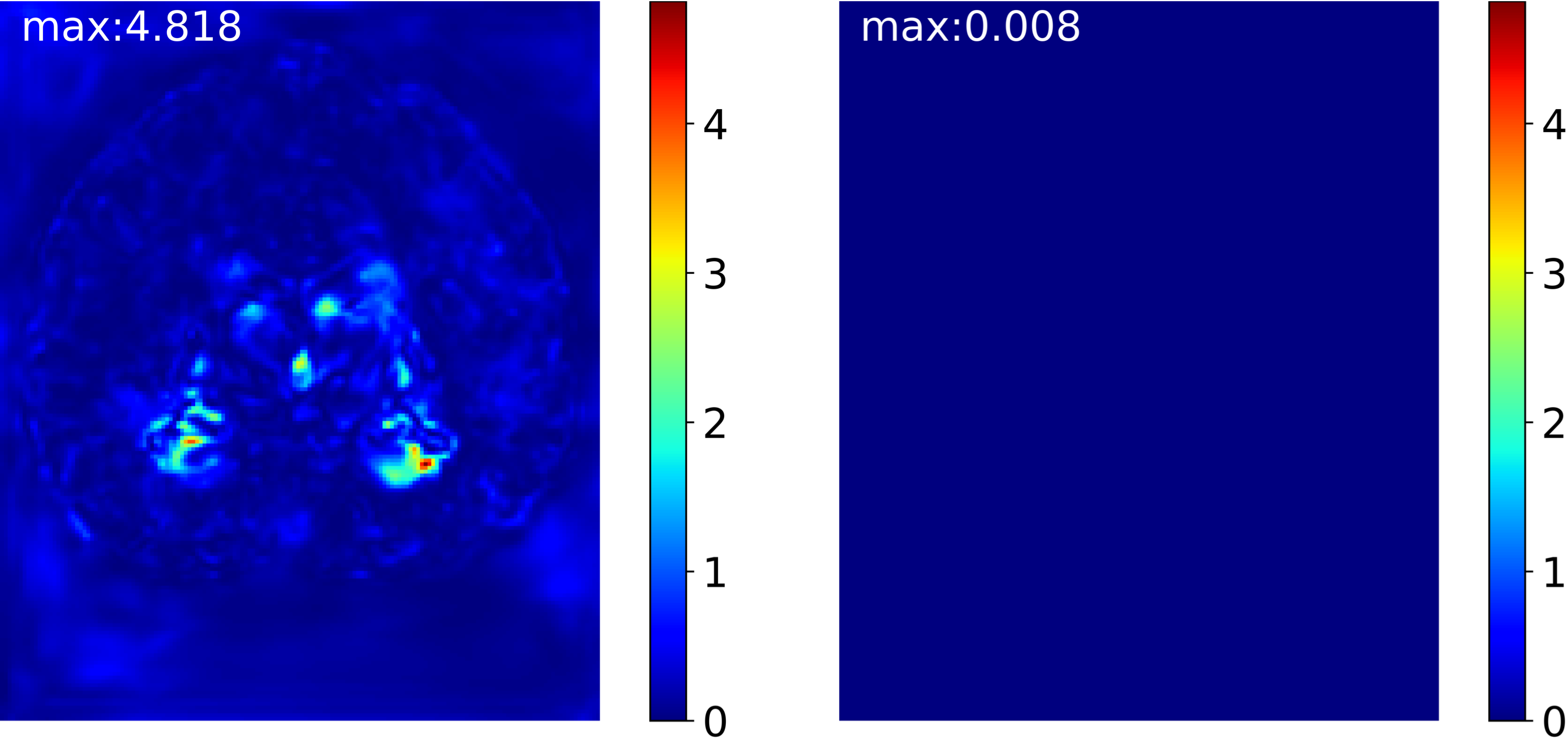}
    \caption{Self-sanity error maps comparison. \textit{Left}: with no self-sanity check, \textit{Right}: with self-sanity check. 
    We unify the error maps' scale bars for a fair comparison.}
    \label{fig:method_self}
\end{figure}
\subsection{Self-sanity and Cross-sanity Checks}
To increase the discriminative power of identical pairs feeding into the model, we propose our self-sanity check as
\begin{equation}
    g^{a\rightarrow a}= 0, \forall a \in \{m, f\}.
    \label{eq:san_self_sanity}
\end{equation}
where the mapping function $g$ learned using any models is restricted to output zero displacements for identical pairs. Such identical pairs can be filtered out using similarity measurements. However, users are unlikely to perform these filters, especially when they do not know that trained models would produce terrible predictions for identical pairs. Hence, our self-sanity check is a natural remedy.

Next, we enforce the inverse consistency on different mappings for $g$ by the model. We will search for correspondence in the fixed image for every point in the moving image such that the transformations between the two images are inconsistent. For example, suppose that a point in the space of image $a$ is registered to the space of image $b$. If we register this point from image $b$ back to image $a$, the point will arrive at the same location in image $a$ (Def.~\ref{def:inverse}). We first define the backward displacement map as in optical flow studies~\cite{hur2017mirrorflow, meister2018unflow, liu2019selflow}, back-projected from $g^{b\rightarrow a}$
\begin{equation}
    \tilde{g}^{b\rightarrow a}(p) = g^{b\rightarrow a}(p+g^{a\rightarrow b}(p)),
\end{equation}
making it convenient for calculation. We then introduce our cross-sanity check in the form of
\begin{equation}
    \begin{aligned}[b]
        &\snormtwo{g^{a\rightarrow b}+\tilde{g}^{b\rightarrow a}} < \alpha(\snormtwo{g^{a\rightarrow b}} + \snormtwo{\tilde{g}^{b\rightarrow a}})+\beta,
    \end{aligned}
    \label{eq:san_cross_sanity}
\end{equation}
$\forall (a,b) \in \{(m,f), (f,m)\}$. Here, we allow the estimation errors to increase linearly with the displacement magnitude with slope $\alpha$ and intercept $\beta$. Instead of imposing zero-tolerance between forward and back-projected backward displacements~\cite{Greer_2021_ICCV, mok2022unsupervised}, we relax the inverse consistency with error tolerance, defined by $\alpha$ and $\beta$, to allow occlusions which is more practical. I.e., This sanity check states that every point $p$ in the moving image $a$ should be able to map back from the fixed image $b$ to its original place in image $a$ with certain error tolerance. We then prove that this error tolerance is upper bounded.
\begin{theorem}[Relaxed registration via cross-sanity check] 
\label{thm:relaxation}
An ideal symmetric registration meets $\varphi^{a\rightarrow b}\circ\varphi^{b\rightarrow a}=id$, defined in Def.~\ref{def:inverse}. Then, a cross-sanity checked registration is a relaxed solution to this ideal registration, satisfying
\begin{equation}
     \normtwotwo{g^{a\rightarrow b}+\tilde{g}^{b\rightarrow a}} < \frac{\beta(2-\alpha) N}{1-\alpha}.
    \label{eq:san_cs_bound}
\end{equation}
\end{theorem}
Here, $0<\alpha<1$ and $\beta>0$. $N$ is a constant, representing the total pixel/voxel numbers. Theoretically, our proposed cross-sanity checks can be viewed as a relaxed version of the strict symmetric constraint, which is commonly used. We also derive the lower/upper bound for satisfying our forward/backward consistency check as shown in \cref{eq:san_cs_bound}. The derivations' details are shown in the Appx.~\ref{sec:thm_relaxation}. To sum up, our cross-sanity check allows a series of solutions to the forward/backward consistency in an enclosed set with a radius of $\sqrt{\frac{\beta(2-\alpha)N}{1-\alpha}}$.

\begin{figure}[!t]
    \centering
    \includegraphics[width=\linewidth]{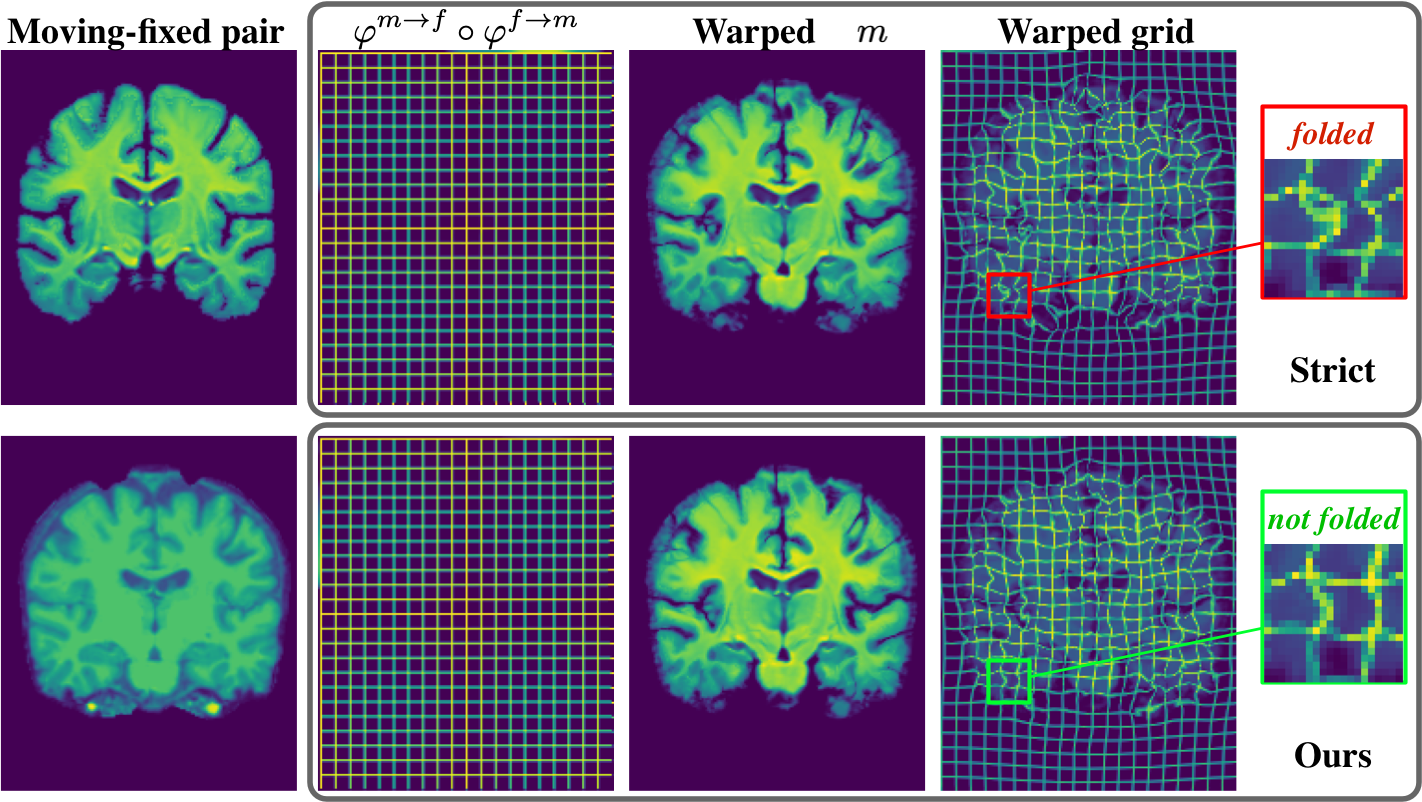}
    \caption{Comparisons between strict inverse consistency trained results (\textit{Top}) and cross-sanity checked results (\textit{Bottom}). Our relaxed sanity-checked result maintains a similar level of inverse consistency as $\varphi^{m\rightarrow f}\circ\varphi^{f\rightarrow m}$ is close to \textit{id} transformation (second column). We can also observe that ours produces a more regular map, compared to the folded map from the model trained with strict inverse consistency. Best view zoomed.}
    \label{fig:method_cross}
\end{figure}
\subsection{Unique Minimizer for Single Image Pair}
Next, we show that there exists a unique solution for our sanity-checked optimization, in terms of a single image pair. We start with writing the standard optimization with our proposed sanity checks in the norm form as
\begin{equation}
\begin{aligned}[b]
    &\quad\,\min -\Sim (f, m \circ (g^{m\rightarrow f}+p)) + \lambda_r\normtwotwo{\Reg(g^{m\rightarrow f})},\\ 
    &\st\normtwotwo{g^{a\rightarrow b}+\tilde{g}^{b\rightarrow a}} < \alpha(\normtwotwo{g^{a\rightarrow b}} + \normtwotwo{\tilde{g}^{b\rightarrow a}})+\beta N,\\ 
    &\quad\quad\,\normtwotwo{g^{a\rightarrow a}}= 0, \quad \forall (a,b) \in \{(m,f), (f,m)\}.
\end{aligned}
\label{eq:san_mini_checked}
\end{equation}
\vspace{-1em}
\begin{theorem}[Existence of the unique minimizer for our relaxed optimization] 
\label{thm:minimizer}
Let $m$ and $f$ be two images defined on the same spatial domain $\Omega$, which is connected, closed, and bounded in $\sR^n$ with a Lipschitz boundary $\partial \Omega$. Let $g: H\times H \rightarrow H^\star$ be a displacement mapping from the Hilbert space $H$ to its dual space $H^\star$. Then, there exists a unique minimizer $g^\star$ to the relaxed minimization problem.
\end{theorem}
The detailed proof for Thm.~\ref{thm:minimizer} is shown in Appx.~\ref{sec:thm_minimizer}. In summary, for a moving-fixed pair, since the relaxation by sanity checks and data term (e.g. SSD) are convex, together with the convex search space for $g$, our problem has a unique minimizer. Thus, the optimization problem is well-conditioned by properties of our regularity constraints such that $g$ will not change dramatically.
\subsection{Loyal Sanity-Checked Minimizer}
In the next section, we further prove that the distance between the minimizer of the sanity-checked constrained optimization and the optimal minimizer for a single moving-fixed image pair can be represented and controlled by our proposed sanity checks. We first rewrite \cref{eq:san_mini_checked} as
\begin{equation}
\begin{aligned}[b]
    &\quad\,\min -\Sim (b, a \circ (g^{a\rightarrow b}+p)) + \lambda_r\normtwotwo{\Reg(g^{a\rightarrow b})},\\
    &\st \normtwotwo{g^{a\rightarrow b}+\tilde{g}^{b\rightarrow a}} < \alpha(\normtwotwo{g^{a\rightarrow b}} + \normtwotwo{\tilde{g}^{b\rightarrow a}})+\beta N,\\ &\quad\quad\,\normtwotwo{g^{a\rightarrow a}}= 0, \quad \forall (a,b) \in \{(m,f), (f,m)\}.
\end{aligned}
\label{eq:san_mini_bichecked}
\end{equation}
\begin{remark}
\vspace{-1em}
We denote this formulation as the bidirectional optimization since it operates in both directions. It is straightforward to show that this bidirectional optimization still satisfies Thm.~\ref{thm:minimizer} so that there exists a unique minimizer.
\end{remark}
Then, with a slight abuse of notations, we define that
\begin{equation}
    \rvg=[\underbrace{g^{m\rightarrow f}, g^{f\rightarrow m}}_{\rvg_c}, \underbrace{g^{m\rightarrow m}, g^{f\rightarrow f}}_{\rvg_s}]^\top.
    \label{eq:san_g_def}
\end{equation}
We can rewrite the minimization problem as
\begin{equation}
\begin{aligned}[b]
    &\min -\Sim (\rvg) + \lambda_r\normtwotwo{\Reg(\rvg)},\\
    \st &\normtwotwo{\rvg_s}= 0,\\
    &\normtwotwo{\rvg_c+\tilde{\rvg_c}} < \alpha(\normtwotwo{\rvg_c} + \normtwotwo{\tilde{\rvg_c}})+2\beta N.
\end{aligned}
\label{eq:san_mini_bichecked_simple}
\end{equation}
Here, $\Sim (\rvg) := \Sim (m, f \circ (g^{f\rightarrow m}+p)) + \Sim (f, m \circ (g^{m\rightarrow f}+p))$, $\normtwotwo{\Reg(\rvg)}:=\normtwotwo{\Reg(g^{m\rightarrow f})} + \normtwotwo{\Reg(g^{f\rightarrow m})}$, $\normtwotwo{\rvg_s}:=\normtwotwo{g^{m\rightarrow m}}+\normtwotwo{g^{f\rightarrow f}}$, $\normtwotwo{\rvg_c+\tilde{\rvg_c}}:=\normtwotwo{g^{m\rightarrow f}+\tilde{g}^{f\rightarrow m}} + \normtwotwo{g^{f\rightarrow m}+\tilde{g}^{m\rightarrow f}}$, and $\normtwotwo{\rvg_c}+\normtwotwo{\tilde{\rvg_c}}:=\normtwotwo{g^{m\rightarrow f}} + \normtwotwo{\tilde{g}^{m\rightarrow f}} + \normtwotwo{g^{f\rightarrow m}} + \normtwotwo{\tilde{g}^{f\rightarrow m}}$. Then, the optimal minimizer $\rvg_\ast$ can be written in the matrix equation as
\begin{equation}
\begin{aligned}[b]
    &\rvg_\ast=\argmin_{\rvg\in H} -\Sim(\rvg), \quad \st \rmA \rvg_\ast=\rvy.\\
    &\rmA = \begin{bmatrix}
        \multicolumn{2}{c}{\Reg(\cdot)} & 0 & 0\\
        \multicolumn{2}{c}{\text{CS}(\cdot)} & 0 & 0\\
        0 & 0 & \multicolumn{2}{c}{\text{SS}(\cdot)} &\\
        \end{bmatrix},
    \quad 
    \rvy = \begin{bmatrix}
        0 \\
        \,\,0^\dag \\
        0 \\
        \end{bmatrix}.
\end{aligned}
\label{eq:san_optimal}
\end{equation}
Here, $\Reg(\cdot):=\normtwotwo{\Reg(\rvg)}$, $\text{CS}(\cdot):=\normtwotwo{\rvg_c+\tilde{\rvg_c}} - \alpha(\normtwotwo{\rvg_c} + \normtwotwo{\tilde{\rvg_c}})-2\beta N$, and $\text{SS}(\cdot):=\normtwotwo{\rvg_s}$, as defined previously. $\dag$ here means that we can have $<0$ solutions for the CS check as in \cref{eq:san_mini_bichecked_simple}, which we show will not interfere with the theoretical foundations by masking in the next section. In this way, we obtain our unique minimizer as
\begin{equation}
\begin{aligned}[b]
    \rvg_{\rm sanity}=\argmin_{\rvg\in H} -\Sim(\rvg) + \frac{\lambda}{2}\normtwotwo{\rmA\rvg-\rvy},
\end{aligned}
\label{eq:san_sanity}
\end{equation}
where $\lambda:=[2\lambda_r, 2\lambda_c, 2\lambda_s]$ is the vector of parameters. 
\begin{theorem}[Loyalty of the sanity-checked minimizer]
\label{thm:loyalty}
Let $\rvg_\ast$ be the optimal minimizer to the bidirectional registration problem, defined in \cref{eq:san_optimal}, and $\rvg_{\rm sanity}$ as our sanity-checked minimizer, defined in \cref{eq:san_sanity}. The distance between these two minimizers can be upper bounded as
\begin{equation}
\begin{aligned}[b]
    \Sim(\rvg_{\rm sanity})-\Sim(\rvg_{\ast})\leq \frac{\lambda}{2}\normtwotwo{A(\rvg_{\rm sanity}-\rvg_{\ast})}.
\end{aligned}
\label{eq:san_distance}
\end{equation}
\end{theorem}
The proof is presented in Appx.~\ref{sec:thm_loyalty}. We then expand the right-hand side of \cref{eq:san_distance} as
\begin{equation}
    \begin{aligned}[b]
        &\Sim(\rvg_{\rm sanity})-\Sim(\rvg_{\ast})\leq\lambda_r\normtwotwo{\Reg(\rvg_{\rm sanity})}\\
        &\quad\qquad\qquad+\lambda_s\normtwotwo{\text{SS}(\rvg_{\rm sanity})})+\lambda_c\normtwotwo{\text{CS}(\rvg_{\rm sanity})},\\
        &\quad\quad\where \Reg(\rvg_\ast) = 0, 
        \, 
        \text{SS}(\rvg_\ast) = 0, 
        \,
        \text{CS}(\rvg_\ast) = 0.
    \end{aligned}
    \label{eq:san_distance_expanded}
\end{equation}
Empirically, the first two terms contribute relatively less (10$\times$ smaller) than the cross-sanity error (See Experiments). Thus, we focus on the CS term here. Following Thm.~\ref{thm:relaxation}, $\normtwotwo{\text{CS}(\rvg_{\rm sanity})}$ is upper bounded in the form of
\begin{equation}
    \begin{aligned}[b]
        \normtwotwo{\text{CS}(\rvg_{\rm sanity})} < 2(1-\alpha)\beta N.
    \end{aligned}
    \label{eq:san_cs_error_bound}
\end{equation}
Here, $0<\alpha<1$ and $\beta>0$ ensure the inequality's direction, where the multiplication of \textit{two} accounts for cross-sanity errors from two directions, i.e., from moving image to fixed image and also fixed image to moving image. We put the full derivation of CS error bound in the Appx.~\ref{sec:prf_cs_bound}. In a nutshell, we prove that by satisfying the cross-sanity check, we have an upper bound for the CS error, constraining the relationship between two displacements from different directions via parameters $\alpha$ and $\beta$.

\begin{lemma}[Upper-bound of distance between optimal minimizer and sanity-checked minimizer]
\label{lem:minimizer}
Let $\rvg_\ast$ and $\rvg_{\rm sanity}$ respectively be the optimal minimizer and the constrained minimizer with cross-sanity check, then we have an upper bound for the distance between such two minimizers as
\begin{equation}
    \Sim(\rvg_{\rm sanity})-\Sim(\rvg_{\ast})< 2\lambda_c(1-\alpha)\beta N.
    \label{eq:san_minimizer_distance}
\end{equation}
\end{lemma}
The Lemma~\ref{lem:minimizer} follows Thm.~\ref{thm:loyalty} and \cref{eq:san_cs_error_bound} (upper bound of CS error). That said, the loyalty w.r.t the distance of the constrained minimizer is controlled by the combination of $\alpha$, $\beta$, and the weight parameter $\lambda_c$.

\textbf{Interpretation of derived upper bound.} By Lemma~\ref{lem:minimizer}, we prove that if we average on the total number of voxels and also two directions, the similarity distance between the optimal minimizer and our sanity-checked minimizer per pixel/voxel is upper bounded by $\lambda_c(1-\alpha)\beta$. That being said, to satisfy our sanity checks, and thus maintain the loyalty to the optimal minimizer, $\lambda_c(1-\alpha)\beta$ should be small. Numerically speaking, for example, if $\alpha=0.1$ and $\beta=10$, we have $(1-\alpha)\beta=9$. This distance of $9$ is extremely large for this delicate image registration task, causing the constrained minimizer to be untrustworthy. Therefore, we need to adopt a relatively small loss weight $\lambda_c$, e.g. $\lambda_c=0.001$, to bound the distance between two minimizers tightly. This observation from our proven theoretical upper bound also coincides with our sanity loss weight ablation study.

\begin{figure}[!t]
    \centering    
    \includegraphics[width=0.95\linewidth]{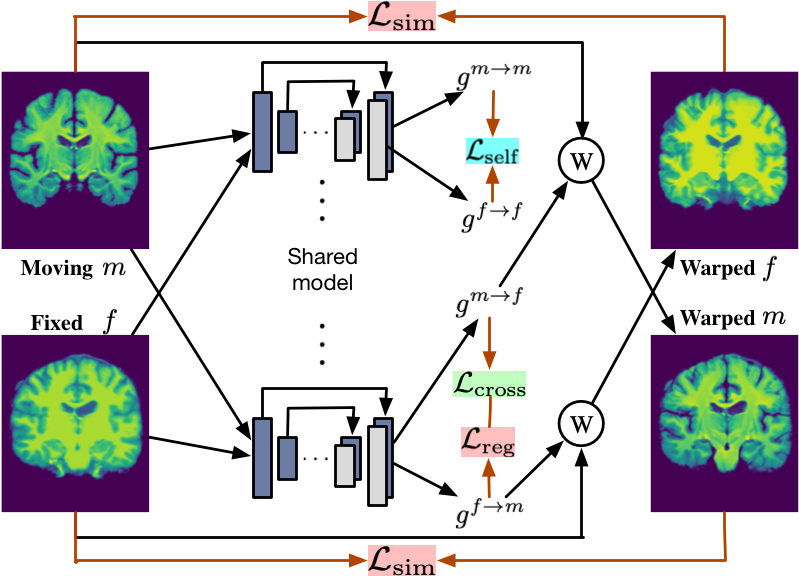}
    \caption{Training a sanity-checked model. \raisebox{1.5pt}{\textcircled{\raisebox{-.9pt} {\scriptsize{W}}}} denotes spatial warping, e.g., warped $m$ is that we warp moving image $m$ using the transformation map calculated from $g^{m\rightarrow f}$. }
    \label{fig:method_pipeline}
\end{figure}
\subsection{Sanity-checked Registration Training}

\begin{figure}[!t]
    \centering
    \includegraphics[width=\linewidth]{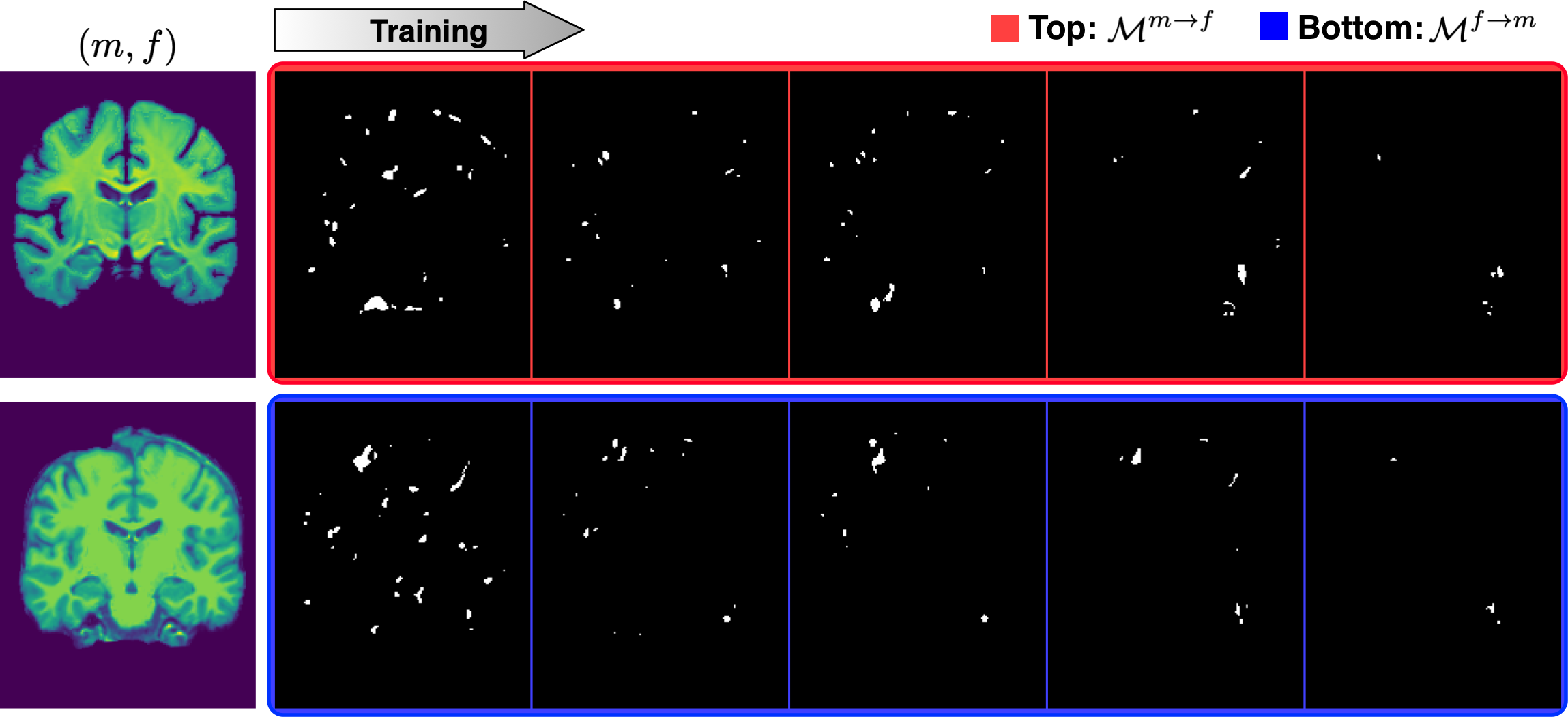}
    \caption{Mask evolution during training. Overall, as training proceeds, violators of the cross-sanity check are decreasing.}
    \label{fig:method_mask}
\end{figure}

We show our sanity-checked training pipeline in~\cref{fig:method_pipeline}. We introduce each loss in the subsection. The self-sanity check can be formulated into a loss function form as 
\begin{equation}
    \mathcal{L}_{\rm self} = \frac{1}{2}(\normtwotwo{g^{m\rightarrow m}} + \normtwotwo{g^{f\rightarrow f}}).
    \label{eq:method_sse}
\end{equation}
So that the self-sanity loss penalizes the squared differences between predicted displacement maps and the ideal ones. Next, we use $m\rightarrow f$ direction as an example ($f\rightarrow m$ direction follows the same principle) to formulate the proposed cross-sanity check loss. We calculate for every voxel and define a binary mask $\mathcal{M}^{m\rightarrow f}$ in the form of
\begin{equation}
    \mathcal{M}^{m\rightarrow f} = 
    \begin{cases}
      0& \text{if satisfies the cross-sanity check},\\
      1& \text{otherwise}.
   \end{cases}
\end{equation}
An interpretation of this binary mask $\mathcal{M}^{m\rightarrow f}$ is that it records violations of the cross-sanity check (\cref{eq:san_cross_sanity}) for each individual point. In this way, $<0$ solution in \cref{eq:san_optimal} will not challenge the theoretical formulations since these points are masked out. Thus, we can formulate the proposed cross-sanity check in the form of a loss function of
\begin{equation}
\begin{aligned}[b]
    \mathcal{L}^{m\rightarrow f}_{\rm cross}&=\normtwotwo{\mathcal{M}^{m\rightarrow f} \odot (g^{m\rightarrow f}+\tilde{g}^{f\rightarrow m})} -\beta\normtwotwo{\mathcal{M}^{m\rightarrow f}}\\
    &- \alpha(\normtwotwo{\mathcal{M}^{m\rightarrow f} \odot g^{m\rightarrow f}} + \normtwotwo{\mathcal{M} ^{m\rightarrow f}\odot \tilde{g}^{f\rightarrow m}}).
\end{aligned}
\label{eq:method_cse}
\end{equation}
The final $\mathcal{L}_{\rm cross}=\frac{1}{2}(\mathcal{L}^{m\rightarrow f}_{\rm cross} + \mathcal{L}^{f\rightarrow m}_{\rm cross})$. Here, $\odot$ denotes element-wise multiplication so that we only retain points violating the cross-sanity check. In this case, the loss value is only calculated for those violators, visualized as occlusion masks, shown in \cref{fig:method_mask}. Finally, the total loss is
\begin{equation}
    \mathcal{L}_{\rm total} = \mathcal{L}_{\rm sim} + \lambda_{r}\mathcal{L}_{\rm reg} + \lambda_{s}\mathcal{L}_{\rm self} + \lambda_{c}\mathcal{L}_{\rm cross}.
    \label{eq:total_loss}
\end{equation}
Here, $\mathcal{L}_{\rm sim}$ as NCC loss, $\mathcal{L}_{\rm reg}$ as $\normtwotwo{\nabla(u)}$, and $\lambda_r=1$, following standard deep registrations~\cite{balakrishnan2019voxelmorph, chen2022transmorph}. If not specified otherwise, $\lambda_{s}=0.1$ and $\lambda_{c}=0.001$. While training, the model optimizes the total loss $\mathcal{L}_{\rm total}$ on different moving-fixed image pairs $(m,f)$ in the training set $\mathcal{D}$ as
\begin{equation}
\begin{aligned}[b]
    &\min_{(m,f)\in\mathcal{D}}\min_{g\in H} \mathcal{L}_{\rm total}.
\end{aligned}
\label{eq:san_mini_model}
\end{equation}
In this way, we fulfill our novel regularization-based sanity-enforcer formulation for training a registration model.
\section{Experiments}
\textbf{Evaluation metrics.} We study model behaviors in a wide range of metrics. For the main metric, we use dice to measure how fit is the transformed segmentation to its ground truth, following previous studies. To study the model taking an identical pair as inputs, we also report self dice (SDice), i.e., the dice when registering the moving image to itself. For pre-operative and post-recurrence registration, we measure the mean target registration error (TRE) of the paired landmarks with Euclidean distance in millimeters and also self mean registration error (STRE) to study the self-sanity of the models. Besides, we report 95\% Hausdorff Distance (HD95) as in~\cite{chen2022transmorph}, which measures the 95th percentile of the distances between boundary points of the transformed subject and the actual subject. We follow~\cite{baheti2021brain, mok2022unsupervised} to report robustness (ROB) for a pair of scans as the relative number of successfully registered landmarks.

For diffeomorphism measurements, we report the percentage of folded voxels (FV) whose Jacobian determinant $<0$, the Absolute Value of Negative Jacobian (AJ) where we sum up all negative Jacobian determinants, and the Standard Deviation of the logarithm of the Jacobian determinant (SDlogJ). All these three Jacobian determinant-related metrics reflect how regular are the transformation maps.

For quantifying sanity errors, we present the mean of self-sanity error (SSE) and the mean of cross-sanity error (CSE), defined in~\cref{eq:method_sse} and \cref{eq:method_cse}, respectively. These two metrics are designed to study model behaviors per the level of each single image pair and are essential for our sanity analysis of different learning-based deep models.

\begin{figure}[!t]
    \centering
    \includegraphics[width=0.95\linewidth]{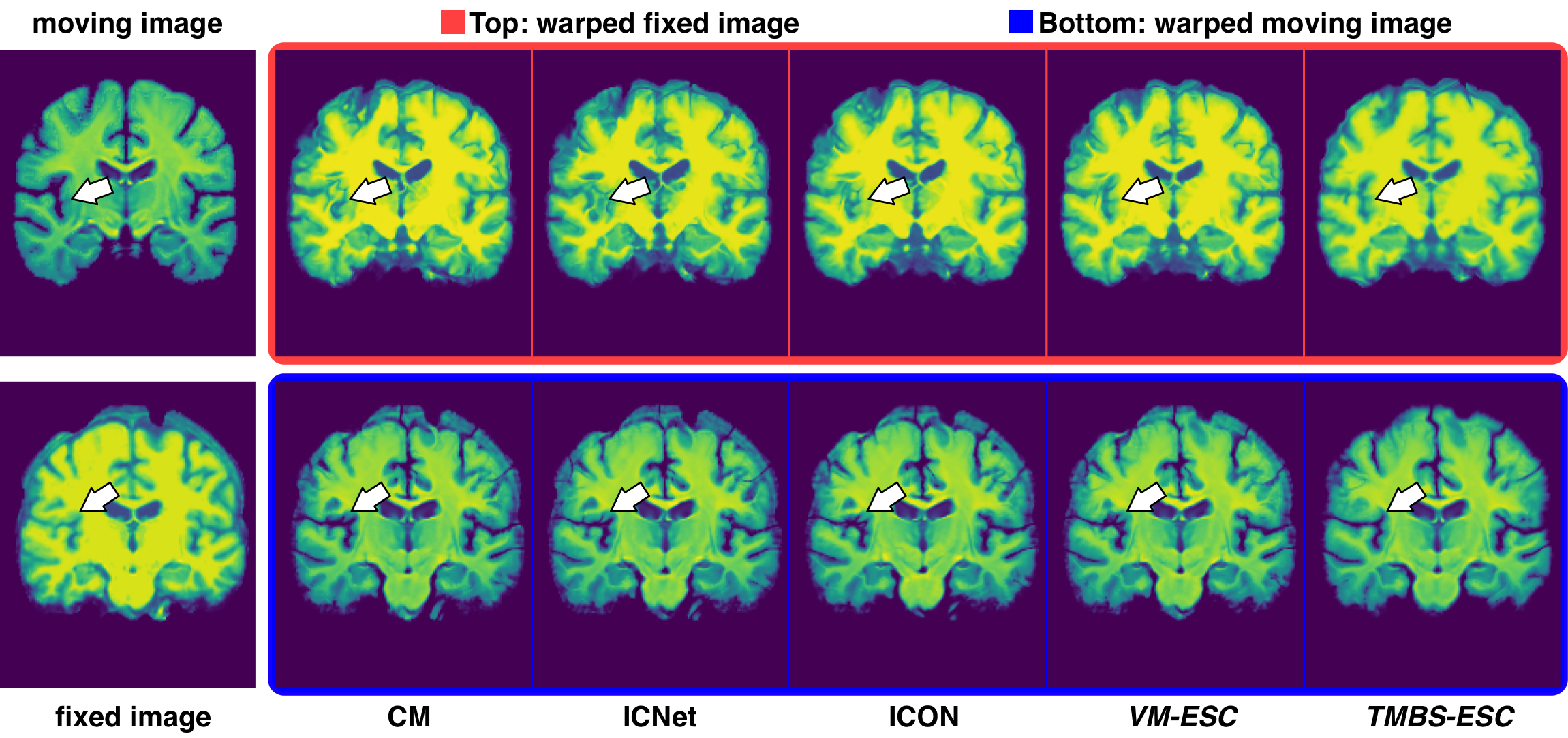}
    \caption{Comparisons between different models on IXI dataset.}
    \label{fig:all}
\end{figure}

\textbf{Implemented models.} We denote the bidirectional optimization as Enclosed (\textit{E}) Image registration, operating to maximize the similarity score and minimize the spatial gradient regularization. We also have Self-sanity checked (\textit{S}) and Cross-sanity checked (\textit{C}) image registrations. \textit{Moreover, as proof of concept that our sanity checks can be applied to different learning-based models}, we implement our proposed techniques on models such as VoxelMorph~\cite{balakrishnan2019voxelmorph} (VM), TransMorph-Large~\cite{chen2022transmorph} (TM), TransMorph-B-spline~\cite{chen2022transmorph} (TMBS), and DIRAC~\cite{mok2022unsupervised}.
\begin{figure}[!t]
    \centering
    \includegraphics[width=0.99\linewidth]{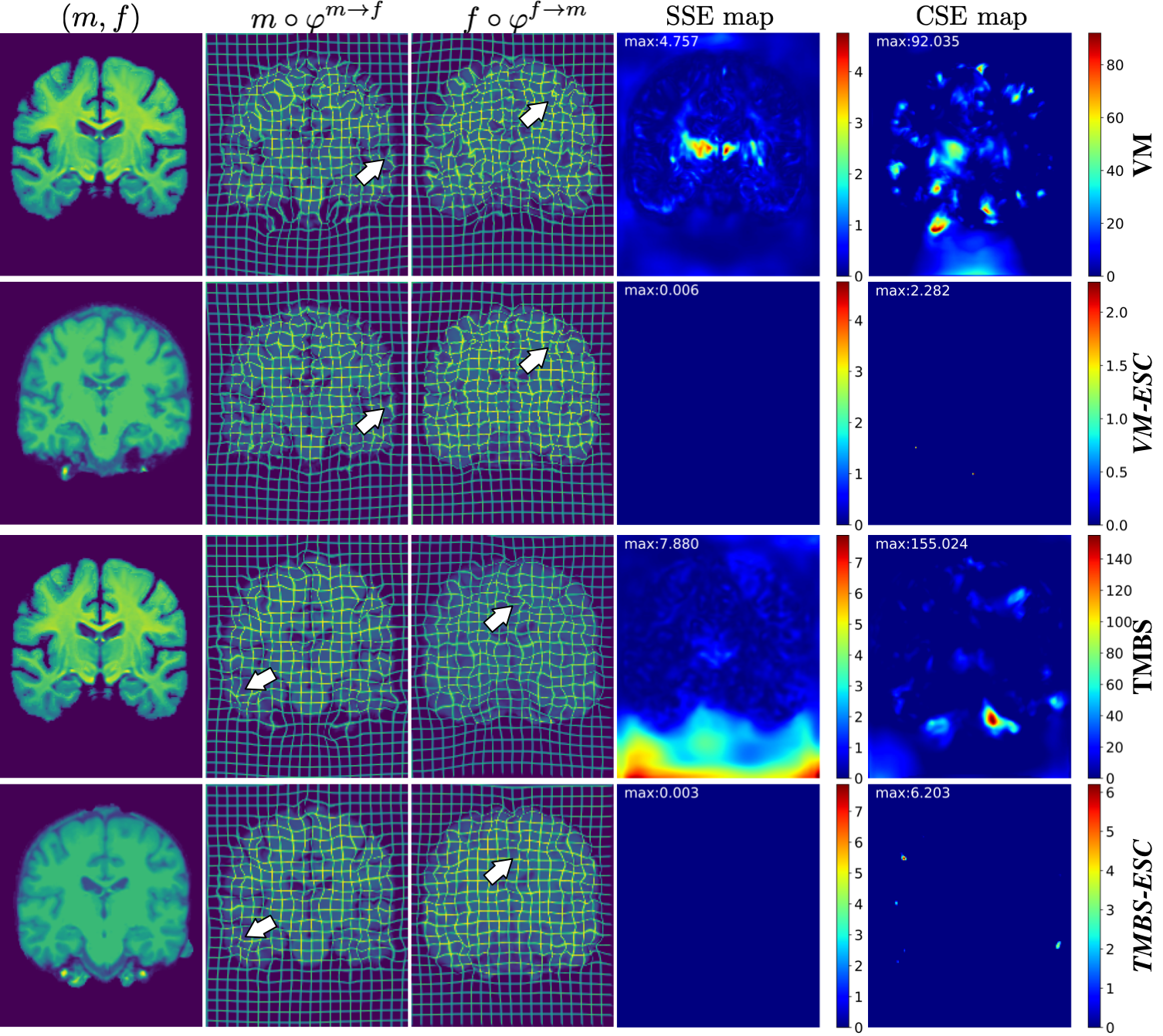}
    \caption{Qualitative comparisons on IXI, where we mark maximum values of error maps on each top left. Best view zoomed.}
    \label{fig:IXI}
\end{figure}
\begin{figure}[!t]
    \centering
    \includegraphics[width=0.99\linewidth]{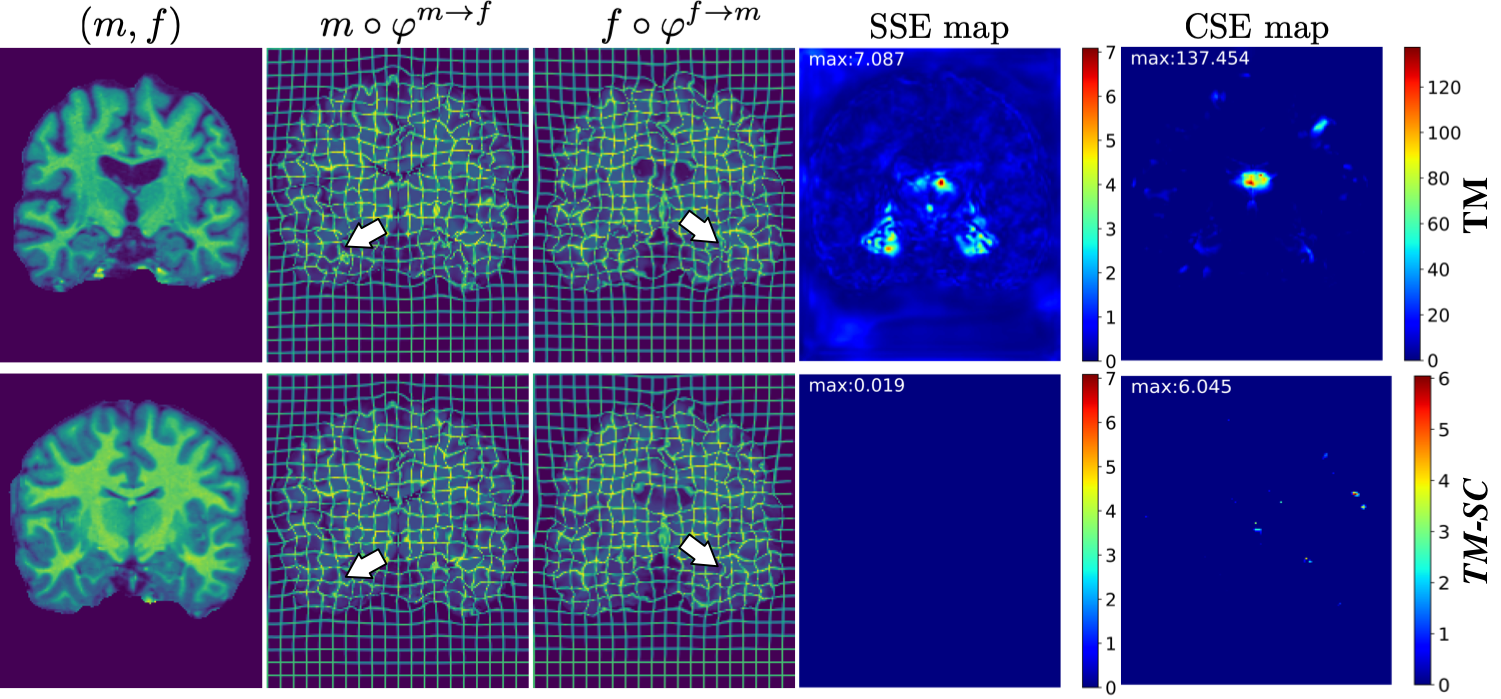}
    \caption{Comparisons on OASIS validation set. Our method produces more regular maps for all input image pairs.}
    \label{fig:OASIS}
\end{figure}
\subsection{Results}
\begin{table*}[!t]
    \begin{minipage}{.3\textwidth}\small
        \centering
        \scalebox{0.85}{\begin{tabular}{lcc}
        \toprule
        Method & Dice\textuparrow & FV\textdownarrow\\
        \midrule
        Affine & 0.386$\pm$0.195 & -\\
        SyN \cite{avants2008symmetric} & 0.645$\pm$0.152 & $<$0.01\\
        NiftyReg \cite{modat2010fast} & 0.645$\pm$0.167 & 0.020$\pm$0.046\\
        LDDMM \cite{beg2005computing} & 0.680$\pm$0.135 & $<$0.01\\
        deedsBCV \cite{heinrich2015multi} & 0.733$\pm$0.126 & 0.147$\pm$0.050\\
        MIDIR \cite{qiu2021learning} & 0.742$\pm$0.128 & $<$0.01\\
        PVT \cite{wang2021pyramid} & 0.727$\pm$0.128 & 1.858$\pm$0.314\\
        CoTr \cite{xie2021cotr} & 0.735$\pm$0.135 & 1.298$\pm$0.342\\
        VM \cite{balakrishnan2019voxelmorph} & 0.732$\pm$0.123 & 1.522$\pm$0.336\\
        VM-diff \cite{dalca2019unsupervised} & 0.580$\pm$0.165 & $<$0.01\\
        ICNet~\cite{zhang2018inverse} & 0.742$\pm$0.223 & 0.854$\pm$0.316\\
        CM \cite{kim2021cyclemorph} & 0.737$\pm$0.123 & 1.719$\pm$0.382\\
        ViT-V-Net \cite{chen2021vit} & 0.734$\pm$0.124 & 1.609$\pm$0.319\\
        nnFormer \cite{zhou2021nnformer} & 0.747$\pm$0.135 & 1.595$\pm$0.358\\
        ICON \cite{Greer_2021_ICCV}& 0.752$\pm$0.155 & 0.695$\pm$0.248\\
        TM \cite{chen2022transmorph} & 0.754$\pm$0.124 & 1.579$\pm$0.328\\
        TM-diff \cite{chen2022transmorph} & 0.594$\pm$0.163 & $<$0.01\\
        TMBS \cite{chen2022transmorph} & 0.761$\pm$0.122 & $<$0.01\\
        \midrule
        \textit{VM-ESC} & 0.743$\pm$0.025 &0.478$\pm$0.101\\
        \textit{TMBS-ESC} & 0.762$\pm$0.023 & $<$0.01\\
        \bottomrule
        \end{tabular}}
        \caption{IXI dataset results.}
        \label{tab:IXI}
    \end{minipage}
    \hfill
    \begin{minipage}{0.68\textwidth}\small
        \begin{minipage}{1.0\textwidth}\small
        \begin{subtable}[t]{0.6\textwidth}\small
            \centering
            \scalebox{0.85}{\begin{tabular}{lcccc}
            \toprule
            Method & Dice\textuparrow & HD95\textdownarrow & SDlogJ\textdownarrow\\
            \midrule
            VTN \cite{lv2022joint} & 0.827$\pm$0.013 & 1.722$\pm$0.318&0.121$\pm$0.015 \\
            ConAdam \cite{siebert2021fast} & 0.846$\pm$0.016& 1.500$\pm$0.304& 0.067$\pm$0.005 \\
            VM \cite{balakrishnan2019voxelmorph} & 0.847$\pm$0.014 &1.546$\pm$0.306 &0.133$\pm$0.021\\
            ClapIRN \cite{mok2021conditional} & 0.861$\pm$0.015& 1.514$\pm$0.337& 0.072$\pm$0.007 \\
            TM \cite{chen2022transmorph} & 0.862$\pm$0.014& 1.431$\pm$0.282& 0.128$\pm$0.021\\
            \midrule
            \textit{TM-SC} & 0.862$\pm$0.014& 1.449$\pm$0.310&0.084$\pm$0.005\\
            \bottomrule
            \end{tabular}}
            \caption{OASIS validation set results.}
            \label{tab:OASIS_val}
        \end{subtable} %
        \hfill
        \begin{subtable}[t]{0.4\textwidth}\small
            \centering
            \scalebox{0.85}{\begin{tabular}{lcccc}
            \toprule
            Method & Dice\textuparrow & HD95\textdownarrow & SDlogJ\textdownarrow\\
            \midrule
            Initial & 0.56 & 3.86& 1.50 \\
            VTN \cite{lv2022joint} & 0.80 & 1.77&0.08 \\
            ConAdam \cite{siebert2021fast} & 0.81& 1.63& 0.07 \\
            ClapIRN \cite{mok2021conditional} & 0.82& 1.67& 0.07 \\
            TM \cite{chen2022transmorph} & 0.82 & 1.66 & 0.12 \\
            \midrule
            \textit{TM-SC} & 0.820& 1.666& 0.085\\
            \bottomrule
            \end{tabular}}
            \caption{OASIS test set results.}
            \label{tab:OASIS_test}
        \end{subtable}
        \caption{OASIS dataset results, obtained from the \textsc{learn2reg} challenge organizers~\cite{hering2022learn2reg}. Since the listed top-ranking methods are already enclosed, we mainly study the impact of adding sanity checks, where the impact is proved to be negligible on these main metrics.}
        \vspace{0.3cm}
        \label{tab:OASIS}
        \end{minipage}
        \begin{minipage}{1.0\textwidth}\small
        \centering
        \scalebox{0.85}{\begin{tabular}{lccccc}
            \toprule
            Method & TRE\textdownarrow & STRE\textdownarrow & ROB\textuparrow& FV\textdownarrow & AJ\texttimes{10}\textsuperscript{2}\textdownarrow\\
            \midrule
            Initial & 6.864$\pm$0.996& -& - &- &-\\
            \midrule
            $^\ast$DIRAC \cite{mok2022unsupervised} & 2.760$\pm$0.247&0.274$\pm$0.027& 0.776$\pm$0.055& 0.025$\pm$0.009&4.242$\pm$2.954 \\
            $^\dag$\textit{DIRAC-SC} &2.719$\pm$0.259&0.218$\pm$0.046 & 0.795$\pm$0.034& 0.022$\pm$0.005&3.001$\pm$1.314 \\
            \bottomrule
            \end{tabular}}
            \caption{BraTSReg dataset results, reported as means$\pm$deviations out of five-fold validations. $\ast$: We use their code and train models until convergence. $\dag$: We implement the cross-sanity check to replace the inverse consistent error part of DIRAC, and add our self-sanity check.}
            \label{tab:dirac}
        \end{minipage}
    \end{minipage}
    \vspace{-0.3cm}
\end{table*}
\textbf{Atlas-to-subject registration.} We split in total 576 T1–weighted brain MRI images from the Information eXtraction from Images (IXI)~\cite{IXI} database into 403, 58, and 115 volumes for training, validation, and test sets. The moving image is an atlas brain MRI obtained from~\cite{kim2021cyclemorph}. FreeSurfer~\cite{fischl2012freesurfer} was used to pre-process the MRI volumes. The pre-processed image volumes are all cropped to the size of 160\texttimes192\texttimes224. Label maps, including 30 anatomical structures, are obtained using FreeSurfer for evaluating registration. Besides CNN-based models~\cite{balakrishnan2019voxelmorph, kim2021cyclemorph, chen2021vit}, we also include other transformer-based deep methods~\cite{chen2021vit, zhou2021nnformer, xie2021cotr, wang2021pyramid} as baselines following~\cite{chen2022transmorph}. The comparison results are presented in~\cref{tab:IXI}, and qualitative comparisons in \cref{fig:all}. For cross-sanity check, we set $\alpha=0.1$ and $\beta=12$. Overall, we can observe models with sanity checks achieve better diffeomorphisms without impairing any Dice performance and even improve the performance for the VM model. We show qualitative comparisons in \cref{fig:IXI}. Compared to the naive counterparts, our sanity-checked model produces more regular maps from both directions and reduces huge sanity errors, comparably using error heatmaps for both self-sanity and cross-sanity errors.

\textbf{Subject-to-subject registration.} OASIS dataset~\cite{marcus2007open, hoopes2022learning} contains a total of 451 brain T1 weighted MRI images, with 394/19/38 images used for training/validation/testing purposes, respectively. The pre-processed image volumes are all cropped to the size of 160\texttimes192\texttimes224. Label maps for 35 anatomical structures are provided using FreeSurfer~\cite{fischl2012freesurfer} for evaluation. For the cross-sanity check, we set $\alpha=0.1$ and $\beta=10$. The results are shown in \cref{tab:OASIS}. In terms of main metrics, we achieve on-par performance with all the state-of-the-art methods, further certifying our sanity checks will not contaminate the performance of models. We also show our qualitative comparison in \cref{fig:OASIS}, indicating that our sanity-enforced image registration results in less self-sanity error from the self-sanity error map comparison and more regular maps in the deformed grid. All these comparisons prove the efficacy of our regularization-based sanity-enforcer to prevent both sanity errors.

\begin{figure}[!t]
    \centering
    \includegraphics[width=0.99\linewidth]{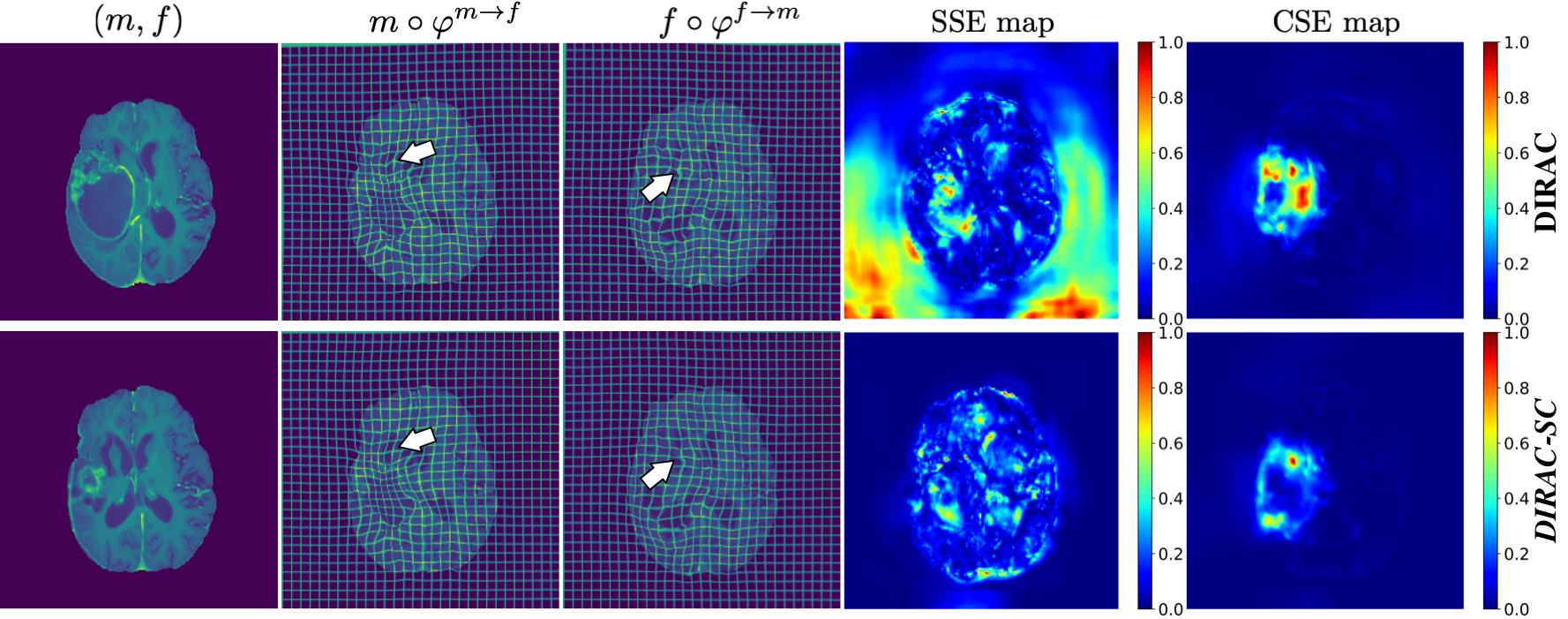}
    \caption{Example axial T1-weighted MR slices comparisons on BraTSReg dataset. The error maps are normalized to incorporate displacements in unit space predicted by DIRAC.}
    \label{fig:brast}
\end{figure}
\textbf{Pre-operative and post-recurrence scans registration.} BraTSReg~\cite{baheti2021brain} dataset contains 160 pairs of pre-operative and follow-up brain MR scans of glioma patients taken from different time points. Each time point contains native T1, contrast-enhanced T1-weighted, T2-weighted, and FLAIR MRI. Within the dataset, 140 pairs of scans are associated with 6 to 50 manual landmarks in both scans. The other 20 pairs of scans only have landmarks in the follow-up scan. Following~\cite{mok2022unsupervised}, we create a five-fold schema where each fold has 122/10/28 for training/validation/test, respectively. We set $\alpha=0.01$ and $\beta=0.03$ in the cross-sanity check since DIRAC predicts normalized displacement (range 0 to 1). We take an average of five folds to report our results in \cref{tab:dirac}. All baselines and our models use the same training setup. We specifically focus on the model training part and directly train all models on data downloaded from the official website~\cite{Bras} without pre-affine and post-processing steps mentioned in their paper~\cite{mok2022robust}. The best model is selected on the validation set, which has the lowest TREs. Despite the high performance of the baseline, we can observe that the substitution by cross-sanity check and the addition of self-sanity check enable all metrics' improvements to be claimed as state-of-the-art. We also show qualitative comparisons in \cref{fig:brast}, where the comparison between SSE error maps demonstrates the effectiveness of our self-sanity check.

\subsection{Sanity Analysis}
We conduct a broad sanity analysis on a wide range of models on the IXI dataset, including popular models (VM~\cite{balakrishnan2019voxelmorph} and TM~\cite{chen2022transmorph}), model with cycle consistency (CM~\cite{kim2021cyclemorph}), models with inverse consistency (ICNet~\cite{zhang2018inverse} and ICON~\cite{Greer_2021_ICCV}), probabilistically-formulated models (VM-diff~\cite{dalca2019unsupervised} and TM-diff~\cite{chen2022transmorph}), and models with diffeomorphic add-ons (MIDIR~\cite{qiu2021learning} and TMBS~\cite{chen2022transmorph}). The results are shown in \cref{tab:sanity_analysis}. We find that models with probabilistic formulation or cycle consistency have higher self-sanity, whereas the other models are insensitive to this scenario and have an inferior SDice. However, these probabilistic formulated models suffer from an relatively low Dice score, refraining from practical usage. For CM, albeit low self-sanity error, the correspondence error between image pairs is not negligible. We also find it interesting that models with diffeomorphic add-ons still suffer from large sanity errors, despite their ability to produce a diffeomorphic/regular map, caused by not modeling different mappings from both directions. Overall, our behavior studies show that most models suffer from inferior Dice/SDice performance, lower diffeomorphism, or significant sanity errors. Our proposed sanity checks improve performance from every aspect, producing a more regular map and preventing sanity errors.

\begin{table}[!t]\small
    \centering
    \scalebox{0.83}{\begin{tabular}{lcccccc}
    \toprule
    \multirow{3}{*}{Method} & \multicolumn{2}{c}{{Main Metric\textuparrow}} &  \multicolumn{2}{c}{{Diffeomorphism\textdownarrow}} & \multicolumn{2}{c}{{Sanity Error\textdownarrow}}  \\
    \cmidrule(r){2-3}\cmidrule(r){4-5}\cmidrule(r){6-7} 
     & Dice & SDice & FV & \text{AJ}\texttimes{10}\textsuperscript{4}& \text{SSE}\texttimes{10}\textsuperscript{-1} & CSE\\
    \midrule
    PVT \cite{wang2021pyramid} & 0.727 & 0.695& 1.858& 3.64& 52.95&19.43\\
    CoTr \cite{xie2021cotr} & 0.735&0.846 & 1.298& 2.10& 6.28&21.37\\
    MIDIR \cite{qiu2021learning} & 0.742& 0.850& $<$0.1& $<$0.1& 6.75&15.75\\
    VM \cite{balakrishnan2019voxelmorph} & 0.732& 0.890 & 1.522& 3.00& 5.16 & 16.55\\
    VM-diff \cite{dalca2019unsupervised} & 0.580& 1.000& $<$0.1& $<$0.1 & 0.13& $<$0.1\\
    ICNet~\cite{zhang2018inverse} & 0.742& 0.849& 0.854& 1.90& 13.23&7.48\\
    CM \cite{kim2021cyclemorph} &0.737 & 1.000 & 1.719& 3.44& 1.18&19.47\\
    VIT-V-Net \cite{chen2021vit} & 0.734& 0.908 & 1.609& 2.97& 14.52&12.98\\
    nnFormer \cite{zhou2021nnformer} & 0.747& 0.799& 1.595& 2.90& 13.31&20.68\\
    ICON~\cite{Greer_2021_ICCV} & 0.752& 0.899& 0.695& 0.90& 7.31&5.68\\
    TM \cite{chen2022transmorph} &0.754 &0.873 & 1.579& 2.75& 85.63&19.62\\
    TM-diff \cite{chen2022transmorph} & 0.594 & 1.000& $<$0.1& $<$0.1 & 0.16& $<$0.1\\
    TMBS \cite{chen2022transmorph} &0.761 &0.903 & $<$0.1& 0.02& 23.84&14.08\\
    \midrule
    \textit{VM-ESC} & 0.743& 1.000 &0.478 &0.43 &$<$0.1 & 2.55\\
    \textit{TMBS-ESC} & 0.762 & 1.000& $<$0.1& $<$0.1& $<$0.1 &2.99 \\
    \bottomrule
    \end{tabular}}
    \caption{Sanity analysis of various models on IXI dataset.}
    \label{tab:sanity_analysis}
\end{table}
\subsection{Ablation Study}
\textbf{Cross-sanity error under strict inverse consistency.}
\begin{table}[!t]\footnotesize
    \centering
    \scalebox{0.8}{\begin{tabular}{cccccccccc|cc}
    \toprule
    \cite{wang2021pyramid} & \cite{xie2021cotr} & \cite{balakrishnan2019voxelmorph} & \cite{kim2021cyclemorph} & \cite{chen2021vit} & \cite{zhou2021nnformer} & \cite{zhang2018inverse}&\cite{Greer_2021_ICCV} & \cite{chen2022transmorph}& \cite{balakrishnan2019voxelmorph}  & $\frac{\textit{TMBS}}{\textit{ESC}}$ & $\frac{\textit{VM}}{\textit{ESC}}$  \\
    \midrule
    19.24&  10.02& 5.63&  7.58&  11.27& 10.44 & 3.46 & 2.78 & 22.61& 7.31& \underline{1.96}& \textbf{0.92}\\
    \bottomrule
    \end{tabular}}
    \caption{Cross-sanity Error using strict inverse consistency on IXI dataset, i.e., $\tilde{g}^{b\rightarrow a}+ g^{a\rightarrow b}=0$. We choose VM for \cite{balakrishnan2019voxelmorph} and TMBS for \cite{chen2022transmorph} as our baselines. We show our results in the last two columns. Still, we achieve state-of-the-art performance even under strict inverse consistency for calculating cross-sanity error.}
    \label{tab:sanity_cice}
\end{table}
We testify the performance under strict inverse consistency as in Def.~\ref{def:inverse}, the same as CICE in~\cite{christensen2006introduction}, shown in~\cref{tab:sanity_cice}. The reductions of strict inverse consistency error compared using relaxed inverse consistency are caused by many small errors taken into account for calculating the average, where our sanity-checked methods still show better performance, compared to all models, including the model trained using explicitly strict inverse consistency, e.g. \cite{Greer_2021_ICCV}.

\textbf{Sanity checks loss weight study.} We derive an upper bound for our saner registration in Lemma~\ref{lem:minimizer}. We show in the study that we can utilize this upper bound to guide us to set appropriate loss weight $\lambda_c$ for the cross-sanity check. For fast verification, we randomly sample 50 subjects from the training set of the IXI dataset and validate/test in the full validation/test set. We also include the self-sanity check loss weight $\lambda_s$ here for comparing purposes. The experimental results are shown in \cref{tab:abs_sanity_weight}. Compared to $\lambda_c$, $\lambda_s$ is insensitive to weight change. As we discussed earlier, if we set higher loss weight to cross-sanity check, for example, 0.01/0.001, approximately having upper bounds as 0.1/0.01, respectively. This creates a huge difference when we train our models. This high tolerance for error results in significantly degenerated performance for Dice, where our derived upper bound gives useful guidance for the setting of loss weights. More discussions can be found in Appendix.

\begin{table}[!t]\small
    \centering
    \scalebox{0.85}{\begin{tabular}{llcccccc}
    \toprule
    $\lambda_c$ & $\lambda_s$ & Dice\textuparrow & SDice\textuparrow & FV\textdownarrow & \text{AJ}\texttimes{10}\textsuperscript{4}\textdownarrow & \text{SSE}\texttimes{10}\textsuperscript{-1}\textdownarrow & CSE\textdownarrow\\
    \cmidrule(r){1-2}\cmidrule(r){3-4}\cmidrule(r){5-6}\cmidrule(r){7-8} 
    \multirow{3}{*}{5\text{e-}4} & 1\text{e-}3& 0.720& 0.957& 0.913& 1.24&0.658 &6.96\\
    & 1\text{e-}2 & 0.720& 0.983& 0.920& 1.27&0.299 &6.77\\
    & 1\text{e-}1 & 0.719& 1.000& 0.896& 1.22&$<$0.1 &6.73\\
    \midrule
    \multirow{3}{*}{1\text{e-}3} & 1\text{e-}3& 0.721 & 0.967& 0.422 & 0.38& 0.572 & 3.43\\
    & 1\text{e-}2 & 0.720& 0.989& 0.426& 0.40&0.301 &3.46\\
    & 1\text{e-}1 &0.721 & 1.000& 0.443& 0.40& $<$0.1&3.43\\
    \midrule
    \multirow{3}{*}{1\text{e-}2} & 1\text{e-}3& 0.473&0.912 &$<$0.1& $<$0.1&1.966 &$<$0.1\\
    & 1\text{e-}2 & 0.420& 0.980& $<$0.1& $<$0.1&1.210&$<$0.1\\
    & 1\text{e-}1 & 0.406& 1.000& $<$0.1& $<$0.1& 0.209&$<$0.1\\
    \bottomrule
    \end{tabular}}
    \caption{Sanity loss weight study of VM on sub-IXI dataset.}
    \label{tab:abs_sanity_weight}
\end{table}
\begin{table}[t]\small
    \centering
    \scalebox{0.85}{\begin{tabular}{lcccccc}
    \toprule
    Method & Dice\textuparrow & SDice\textuparrow & FV\textdownarrow & \text{AJ}\texttimes{10}\textsuperscript{4}\textdownarrow & \text{SSE}\texttimes{10}\textsuperscript{-1}\textdownarrow & CSE\textdownarrow\\
    \cmidrule(r){1-1}\cmidrule(r){2-3}\cmidrule(r){4-5}\cmidrule(r){6-7} 
    VM & 0.732& 0.890 & 1.522& 3.00& 5.156 & 16.55\\
    VM-E & 0.742& 0.919& 1.574& 3.04& 2.647& 26.97\\
    VM-ES & 0.740& 1.000& 1.442& 2.72& 0.050& 25.32\\
    VM-EC & 0.743& 0.950& 0.447& 0.39& 1.178& 2.61\\
    {\textit{VM-ESC}} & 0.743& 1.000& 0.478& 0.43& $<$0.1& 2.55\\
    \midrule
    TM & 0.862& 0.925& 0.752& 1.52& 3.069& 10.72\\
    TM-S & 0.861& 1.000& 0.777& 1.66& 0.018&11.43\\
    TM-C & 0.862& 0.948& 0.246& 0.26&0.991 &2.93\\
    {\textit{TM-SC}} & 0.862& 1.000& 0.307& 0.35& $<$0.1&3.17\\
    \bottomrule
    \end{tabular}}
    \caption{Model ablations on IXI dataset (\textbf{top rows}) and OASIS validation dataset (\textbf{bottom rows}). Since TM is already enclosed (E), we focus on studying the impacts of adding our sanity checks.}
    \label{tab:abs_model}
\end{table}
\textbf{Model ablation study.} We study our proposed sanity checks on IXI and OASIS validation datasets and report the results in~\cref{tab:abs_model}. We show that each sanity check reduces the corresponding errors without compromising other metrics' performance compared to their naive counterparts. We also find that the bidirectional registration and cross-sanity check can also mitigate self-sanity errors to a certain level, but cannot eliminate such errors completely, proving that our self-sanity check is necessary to regulate the models' behavior for mapping identical image pairs. More experiments such as parameters $\alpha$ and $\beta$ numerical study, sanity preservation study, statistical significance of results, ablative qualitative results, etc, can be seen in the Appendix.

\section{Conclusion}
This paper focuses on correcting learning-based deep models' behaviors on single moving-fixed image pairs. In our model sanity analysis, we find that most existing models suffer from significant sanity errors, with no exceptions for models equipped with diffeomorphic add-ons. We show that this sanity-checked model can prevent such sanity errors without contaminating any registration performance. While the experimental results certify the effectiveness of our proposed sanity checks, our sanity checks are supported by a set of theoretical guarantees derived in this paper. We first show that there is an error upper bound for our sanity-checked formulation to the optimal condition. Then, we show that this upper bound can give significant guidance to train a sanity-checked registration model, where we believe it is beneficial for preventing overly optimization on image similarities when training deep registration models.

\section*{Acknowledgements}
Authors thank anonymous reviewers for their constructive comments and help from Yuzhang Shang for proofreading the rebuttal. Authors would also want to thank Junyu Chen, Tony C. W. Mok for their generous help with their papers and codes, and \textsc{learn2reg} challenge organizer for evaluating our models. This work was supported by NSF SCH-2123521. This article solely reflects the opinions and conclusions of its authors and not the funding agency.

{\small
\bibliographystyle{ieee_fullname}
\bibliography{egbib}

\begin{thebibliography}{10}\itemsep=-1pt

\bibitem{IXI}
Ixi brain, 2022.
\newblock \url{https://brain-development.org/ixi-dataset}.

\bibitem{avants2008symmetric}
Brian~B Avants, Charles~L Epstein, Murray Grossman, and James~C Gee.
\newblock Symmetric diffeomorphic image registration with cross-correlation:
  evaluating automated labeling of elderly and neurodegenerative brain.
\newblock {\em MIA}, 12(1):26--41, 2008.

\bibitem{baheti2021brain}
Bhakti Baheti, Diana Waldmannstetter, Satrajit Chakrabarty, Hamed Akbari,
  Michel Bilello, Benedikt Wiestler, Julian Schwarting, Evan Calabrese, Jeffrey
  Rudie, Syed Abidi, et~al.
\newblock The brain tumor sequence registration challenge: Establishing
  correspondence between pre-operative and follow-up mri scans of diffuse
  glioma patients.
\newblock {\em arXiv:2112.06979}, 2021.

\bibitem{bajcsy1989multiresolution}
Ruzena Bajcsy and Stane Kova{\v{c}}i{\v{c}}.
\newblock Multiresolution elastic matching.
\newblock {\em CVGIP}, 46(1):1--21, 1989.

\bibitem{Bras}
Spyridon Bakas.
\newblock Brain tumor sequence registration challenge, 2022.
\newblock \url{https://www.med.upenn.edu/cbica/brats-reg-challenge}.

\bibitem{balakrishnan2019voxelmorph}
Guha Balakrishnan, Amy Zhao, Mert~R Sabuncu, John Guttag, and Adrian~V Dalca.
\newblock Voxelmorph: a learning framework for deformable medical image
  registration.
\newblock {\em TMI}, 38(8):1788--1800, 2019.

\bibitem{beg2005computing}
M~Faisal Beg, Michael~I Miller, Alain Trouv{\'e}, and Laurent Younes.
\newblock Computing large deformation metric mappings via geodesic flows of
  diffeomorphisms.
\newblock {\em IJCV}, 61(2):139--157, 2005.

\bibitem{cao2018deformable}
Xiaohuan Cao, Jianhua Yang, Jun Zhang, Qian Wang, Pew-Thian Yap, and Dinggang
  Shen.
\newblock Deformable image registration using a cue-aware deep regression
  network.
\newblock {\em TBE}, 65(9):1900--1911, 2018.

\bibitem{chen2022transmorph}
Junyu Chen, Eric~C Frey, Yufan He, William~P Segars, Ye Li, and Yong Du.
\newblock Transmorph: Transformer for unsupervised medical image registration.
\newblock {\em MIA}, page 102615, 2022.

\bibitem{chen2021vit}
Junyu Chen, Yufan He, Eric~C Frey, Ye Li, and Yong Du.
\newblock Vit-v-net: Vision transformer for unsupervised volumetric medical
  image registration.
\newblock {\em arXiv:2104.06468}, 2021.

\bibitem{christensen2006introduction}
Gary~E Christensen, Xiujuan Geng, Jon~G Kuhl, Joel Bruss, Thomas~J Grabowski,
  Imran~A Pirwani, Michael~W Vannier, John~S Allen, and Hanna Damasio.
\newblock Introduction to the non-rigid image registration evaluation project
  (nirep).
\newblock In {\em WBIR}, 2006.

\bibitem{christensen1997volumetric}
Gary~E Christensen, Sarang~C Joshi, and Michael~I Miller.
\newblock Volumetric transformation of brain anatomy.
\newblock {\em TMI}, 16(6):864--877, 1997.

\bibitem{dalca2019unsupervised}
Adrian~V Dalca, Guha Balakrishnan, John Guttag, and Mert~R Sabuncu.
\newblock Unsupervised learning of probabilistic diffeomorphic registration for
  images and surfaces.
\newblock {\em MIA}, 57:226--236, 2019.

\bibitem{fischl2012freesurfer}
Bruce Fischl.
\newblock Freesurfer.
\newblock {\em Neuroimage}, 62(2):774--781, 2012.

\bibitem{Fu_2021_CVPR}
Kexue Fu, Shaolei Liu, Xiaoyuan Luo, and Manning Wang.
\newblock Robust point cloud registration framework based on deep graph
  matching.
\newblock In {\em CVPR}, 2021.

\bibitem{Greer_2021_ICCV}
Hastings Greer, Roland Kwitt, Francois-Xavier Vialard, and Marc Niethammer.
\newblock Icon: Learning regular maps through inverse consistency.
\newblock In {\em ICCV}, 2021.

\bibitem{grzech2022variational}
Daniel Grzech, Mohammad~Farid Azampour, Ben Glocker, Julia Schnabel, Nassir
  Navab, Bernhard Kainz, and Lo{\"\i}c Le~Folgoc.
\newblock A variational bayesian method for similarity learning in non-rigid
  image registration.
\newblock In {\em CVPR}, 2022.

\bibitem{hart2009optimal}
Gabriel~L Hart, Christopher Zach, and Marc Niethammer.
\newblock An optimal control approach for deformable registration.
\newblock In {\em CVPR}, 2009.

\bibitem{heinrich2015multi}
Mattias~P Heinrich, Oskar Maier, and Heinz Handels.
\newblock Multi-modal multi-atlas segmentation using discrete optimisation and
  self-similarities.
\newblock {\em VISCERAL Challenge@ ISBI}, 1390:27, 2015.

\bibitem{hering2022learn2reg}
Alessa Hering, Lasse Hansen, Tony C.~W. Mok, Albert~CS Chung, Hanna Siebert,
  Stephanie H{\"a}ger, Annkristin Lange, Sven Kuckertz, Stefan Heldmann, Wei
  Shao, et~al.
\newblock Learn2reg: comprehensive multi-task medical image registration
  challenge, dataset and evaluation in the era of deep learning.
\newblock {\em TMI}, 2022.

\bibitem{hoopes2022learning}
Andrew Hoopes, Malte Hoffmann, Douglas~N Greve, Bruce Fischl, John Guttag, and
  Adrian~V Dalca.
\newblock Learning the effect of registration hyperparameters with hypermorph.
\newblock {\em MELBA}, 3:1--30, 2022.

\bibitem{hur2017mirrorflow}
Junhwa Hur and Stefan Roth.
\newblock Mirrorflow: Exploiting symmetries in joint optical flow and occlusion
  estimation.
\newblock In {\em ICCV}, 2017.

\bibitem{kim2021cyclemorph}
Boah Kim, Dong~Hwan Kim, Seong~Ho Park, Jieun Kim, June-Goo Lee, and Jong~Chul
  Ye.
\newblock Cyclemorph: cycle consistent unsupervised deformable image
  registration.
\newblock {\em MIA}, 71:102036, 2021.

\bibitem{liu2019selflow}
Pengpeng Liu, Michael Lyu, Irwin King, and Jia Xu.
\newblock Selflow: Self-supervised learning of optical flow.
\newblock In {\em CVPR}, 2019.

\bibitem{lv2022joint}
Jinxin Lv, Zhiwei Wang, Hongkuan Shi, Haobo Zhang, Sheng Wang, Yilang Wang, and
  Qiang Li.
\newblock Joint progressive and coarse-to-fine registration of brain mri via
  deformation field integration and non-rigid feature fusion.
\newblock {\em TMI}, 41(10):2788--2802, 2022.

\bibitem{marcus2007open}
Daniel~S Marcus, Tracy~H Wang, Jamie Parker, John~G Csernansky, John~C Morris,
  and Randy~L Buckner.
\newblock Open access series of imaging studies (oasis): cross-sectional mri
  data in young, middle aged, nondemented, and demented older adults.
\newblock {\em Journal of cognitive neuroscience}, 19(9):1498--1507, 2007.

\bibitem{meister2018unflow}
Simon Meister, Junhwa Hur, and Stefan Roth.
\newblock Unflow: Unsupervised learning of optical flow with a bidirectional
  census loss.
\newblock In {\em AAAI}, 2018.

\bibitem{modat2010fast}
Marc Modat, Gerard~R Ridgway, Zeike~A Taylor, Manja Lehmann, Josephine Barnes,
  David~J Hawkes, Nick~C Fox, and S{\'e}bastien Ourselin.
\newblock Fast free-form deformation using graphics processing units.
\newblock {\em Computer methods and programs in biomedicine}, 98(3):278--284,
  2010.

\bibitem{mok2021conditional}
Tony C.~W. Mok and Albert Chung.
\newblock Conditional deformable image registration with convolutional neural
  network.
\newblock In {\em MICCAI}, 2021.

\bibitem{mok2022robust}
Tony C.~W. Mok and Albert Chung.
\newblock Robust image registration with absent correspondences in
  pre-operative and follow-up brain mri scans of diffuse glioma patients.
\newblock {\em arXiv:2210.11045}, 2022.

\bibitem{mok2022unsupervised}
Tony C.~W. Mok and Albert Chung.
\newblock Unsupervised deformable image registration with absent
  correspondences in pre-operative and post-recurrence brain tumor mri scans.
\newblock In {\em MICCAI}, 2022.

\bibitem{pennec1999understanding}
Xavier Pennec, Pascal Cachier, and Nicholas Ayache.
\newblock Understanding the “demon’s algorithm”: 3d non-rigid
  registration by gradient descent.
\newblock In {\em MICCAI}, 1999.

\bibitem{Qin_2022_CVPR}
Zheng Qin, Hao Yu, Changjian Wang, Yulan Guo, Yuxing Peng, and Kai Xu.
\newblock Geometric transformer for fast and robust point cloud registration.
\newblock In {\em CVPR}, 2022.

\bibitem{qiu2021learning}
Huaqi Qiu, Chen Qin, Andreas Schuh, Kerstin Hammernik, and Daniel Rueckert.
\newblock Learning diffeomorphic and modality-invariant registration using
  b-splines.
\newblock In {\em MIDL}, 2021.

\bibitem{qu2022cross}
Lei Qu, Yuanyuan Li, Peng Xie, Lijuan Liu, Yimin Wang, Jun Wu, Yu Liu, Tao
  Wang, Longfei Li, Kaixuan Guo, et~al.
\newblock Cross-modal coherent registration of whole mouse brains.
\newblock {\em Nature Methods}, 19(1):111--118, 2022.

\bibitem{rueckert1999nonrigid}
Daniel Rueckert, Luke~I Sonoda, Carmel Hayes, Derek~LG Hill, Martin~O Leach,
  and David~J Hawkes.
\newblock Nonrigid registration using free-form deformations: application to
  breast mr images.
\newblock {\em TMI}, 18(8):712--721, 1999.

\bibitem{schmah2013left}
Tanya Schmah, Laurent Risser, and Fran{\c{c}}ois-Xavier Vialard.
\newblock Left-invariant metrics for diffeomorphic image registration with
  spatially-varying regularisation.
\newblock In {\em MICCAI}, 2013.

\bibitem{shen2002hammer}
Dinggang Shen and Christos Davatzikos.
\newblock Hammer: hierarchical attribute matching mechanism for elastic
  registration.
\newblock {\em TMI}, 21(11):1421--1439, 2002.

\bibitem{shen2019networks}
Zhengyang Shen, Xu Han, Zhenlin Xu, and Marc Niethammer.
\newblock Networks for joint affine and non-parametric image registration.
\newblock In {\em CVPR}, 2019.

\bibitem{shen2019region}
Zhengyang Shen, Fran{\c{c}}ois-Xavier Vialard, and Marc Niethammer.
\newblock Region-specific diffeomorphic metric mapping.
\newblock {\em NeurIPS}, 2019.

\bibitem{siebert2021fast}
Hanna Siebert, Lasse Hansen, and Mattias~P Heinrich.
\newblock Fast 3d registration with accurate optimisation and little learning
  for learn2reg 2021.
\newblock In {\em MICCAI}, 2021.

\bibitem{sokooti2017nonrigid}
Hessam Sokooti, Bob De~Vos, Floris Berendsen, Boudewijn~PF Lelieveldt, Ivana
  I{\v{s}}gum, and Marius Staring.
\newblock Nonrigid image registration using multi-scale 3d convolutional neural
  networks.
\newblock In {\em MICCAI}, 2017.

\bibitem{vercauteren2009diffeomorphic}
Tom Vercauteren, Xavier Pennec, Aymeric Perchant, and Nicholas Ayache.
\newblock Diffeomorphic demons: Efficient non-parametric image registration.
\newblock {\em NeuroImage}, 45(1):S61--S72, 2009.

\bibitem{vialard2012diffeomorphic}
Fran{\c{c}}ois-Xavier Vialard, Laurent Risser, Daniel Rueckert, and Colin~J
  Cotter.
\newblock Diffeomorphic 3d image registration via geodesic shooting using an
  efficient adjoint calculation.
\newblock {\em IJCV}, 97:229--241, 2012.

\bibitem{wang2020deepflash}
Jian Wang and Miaomiao Zhang.
\newblock Deepflash: An efficient network for learning-based medical image
  registration.
\newblock In {\em CVPR}, 2020.

\bibitem{wang2021pyramid}
Wenhai Wang, Enze Xie, Xiang Li, Deng-Ping Fan, Kaitao Song, Ding Liang, Tong
  Lu, Ping Luo, and Ling Shao.
\newblock Pyramid vision transformer: A versatile backbone for dense prediction
  without convolutions.
\newblock In {\em ICCV}, 2021.

\bibitem{wang2021bi}
Xuechun Wang, Weilin Zeng, Xiaodan Yang, Yongsheng Zhang, Chunyu Fang, Shaoqun
  Zeng, Yunyun Han, and Peng Fei.
\newblock Bi-channel image registration and deep-learning segmentation (birds)
  for efficient, versatile 3d mapping of mouse brain.
\newblock {\em Elife}, 10:63455, 2021.

\bibitem{xie2021cotr}
Yutong Xie, Jianpeng Zhang, Chunhua Shen, and Yong Xia.
\newblock Cotr: Efficiently bridging cnn and transformer for 3d medical image
  segmentation.
\newblock In {\em MICCAI}, 2021.

\bibitem{yang2017quicksilver}
Xiao Yang, Roland Kwitt, Martin Styner, and Marc Niethammer.
\newblock Quicksilver: Fast predictive image registration--a deep learning
  approach.
\newblock {\em NeuroImage}, 158:378--396, 2017.

\bibitem{zhang2018inverse}
Jun Zhang.
\newblock Inverse-consistent deep networks for unsupervised deformable image
  registration.
\newblock {\em arXiv:1809.03443}, 2018.

\bibitem{zhou2021nnformer}
Hong-Yu Zhou, Jiansen Guo, Yinghao Zhang, Lequan Yu, Liansheng Wang, and Yizhou
  Yu.
\newblock nnformer: Interleaved transformer for volumetric segmentation.
\newblock {\em arXiv:2109.03201}, 2021.

\bibitem{zhou2021deep}
S~Kevin Zhou, Hoang~Ngan Le, Khoa Luu, Hien~V Nguyen, and Nicholas Ayache.
\newblock Deep reinforcement learning in medical imaging: A literature review.
\newblock {\em MIA}, 73:102193, 2021.

\end{thebibliography}
}

\clearpage
\appendix
\setcounter{theorem}{0}
\setcounter{table}{0}
\setcounter{figure}{0}
\setcounter{equation}{0}
\setcounter{page}{1}

\renewcommand{\thetable}{A\arabic{table}} 
\renewcommand{\thefigure}{A\arabic{figure}} 
\renewcommand{\theequation}{A\arabic{equation}} 

\section{Appendix}
\subsection{Comparison Between Diffeomorphic Methods and Inverse Consistent Methods}
\begin{figure}[h]\small
    \centering
    \includegraphics[width=\linewidth]{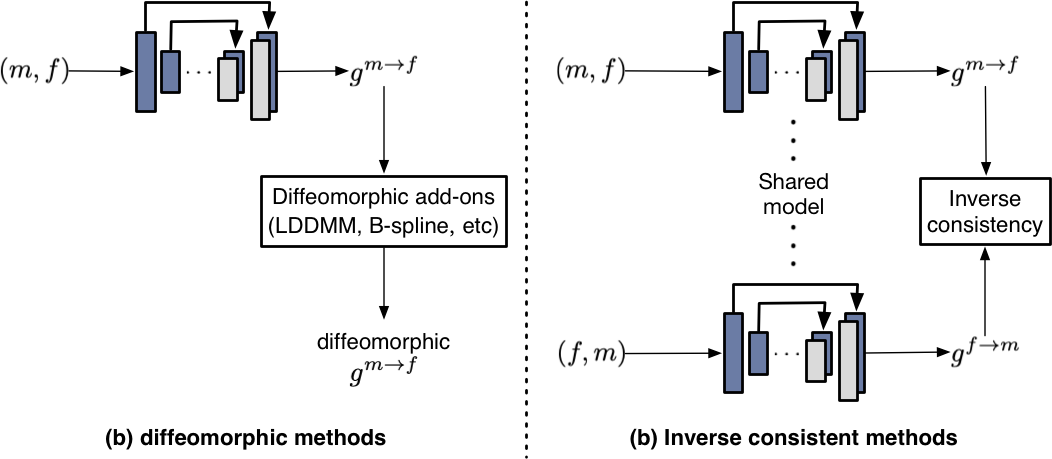}
    \caption{Illustration of two different types of methods. Technically, diffeomorphic methods~\cite{bajcsy1989multiresolution, rueckert1999nonrigid, vialard2012diffeomorphic, schmah2013left, yang2017quicksilver, wang2020deepflash, shen2019networks, shen2019region} are also inverse-consistent. But following the convention in~\cite{zhang2018inverse, Greer_2021_ICCV, mok2022unsupervised}, we restrict inverse consistency in the essence of operating on two different mappings predicted by the same model. Since the diffeomorphic methods only operate on one mapping, where the other mapping is calculated from that mapping, rather than predicted. Thus, they are called diffeomorphic methods instead.}
    \label{fig:mapping}
\end{figure}
\subsection{Proof of Thm.~\ref{thm:relaxation}}
\label{sec:thm_relaxation}
\begin{theorem}[Relaxed ideal symmetric registration via cross-sanity check] 
An ideal symmetric registration meets
\begin{equation*}
    \varphi^{a\rightarrow b}\circ\varphi^{b\rightarrow a}=id,
\end{equation*}
in which $id$ denotes the identity transformation. Then, a cross-sanity checked registration is a relaxed solution to the ideal registration, satisfying
\begin{equation*}
     \normtwotwo{g^{a\rightarrow b}+\tilde{g}^{b\rightarrow a}} <\frac{\beta(2-\alpha) N}{1-\alpha}.
\end{equation*}
\end{theorem}
\begin{prf}
A straightforward explanation for ideal symmetric registration $\varphi^{a\rightarrow b}\circ\varphi^{b\rightarrow a}=id$ would be that the coordinates of one pixel ($p$) for image $a$ stay the same after two transformations: forward transformation (from $a$ to $b$) and backward transformation (from $b$ back to $a$). Therefore, we have the chain of the coordinates changing
\begin{equation}
\begin{aligned}[b]
    p&\\
    &\xRightarrow{a\rightarrow b} g^{a\rightarrow b}(p)+p\\ &\xRightarrow{b\rightarrow a}g^{b\rightarrow a}(g^{a\rightarrow b}(p)+p)+g^{a\rightarrow b}+p\\
    &=p.
\end{aligned}
\end{equation}
By simplification, we have the ideal inverse consistency:
\begin{equation}
\begin{aligned}[b]
    \underbrace{g^{b\rightarrow a}(g^{a\rightarrow b}(p)+p)}&+g^{a\rightarrow b}+p=p\\
    \tilde{g}^{b\rightarrow a}\quad\quad\quad&+\quad g^{a\rightarrow b}\,\,=0.
\end{aligned}
\label{eq:app_strict}
\end{equation}
Different from the ideal inverse consistency, recall that our cross-sanity check in norm form is 
\begin{equation}
    \normtwotwo{g^{a\rightarrow b}+\tilde{g}^{b\rightarrow a}} < \alpha(\normtwotwo{g^{a\rightarrow b}} + \normtwotwo{\tilde{g}^{b\rightarrow a}})+\beta N.
\end{equation}
Suppose $0<\alpha<1$ and $\beta>0$. By expanding the cross-sanity check, we have
\begin{equation}
\begin{aligned}
    {g^{a\rightarrow b}}^\top\tilde{g}^{b\rightarrow a}&<\frac{\beta N-(1-\alpha)(\normtwotwo{g^{a\rightarrow b}}+\normtwotwo{\tilde{g}^{b\rightarrow a}})}{2}\\
    &\leq\frac{\beta N}{2}.
\end{aligned}
\end{equation}
So, we have
\begin{equation}
    \begin{aligned}
        0\leq {g^{a\rightarrow b}}^\top\tilde{g}^{b\rightarrow a}<\frac{\beta N}{2},
    \end{aligned}
\end{equation}
and
\begin{equation}
\begin{aligned}
    0<\normtwotwo{g^{a\rightarrow b}}+\normtwotwo{\tilde{g}^{b\rightarrow a}} < \frac{\beta N}{1-\alpha}.
\end{aligned}
\end{equation}
Thus, we have
\begin{equation}
\begin{aligned}[b]
    \normtwotwo{g^{a\rightarrow b}+\tilde{g}^{b\rightarrow a}} &= \normtwotwo{g^{a\rightarrow b}} + \normtwotwo{\tilde{g}^{b\rightarrow a}} + 2{g^{a\rightarrow b}}^\top\tilde{g}^{b\rightarrow a}\\
    &< \frac{\beta(2-\alpha) N}{1-\alpha}.
\end{aligned}
\end{equation}
Finally, we derive the lower/upper bound as
\begin{equation}
    \normtwotwo{g^{a\rightarrow b}+\tilde{g}^{b\rightarrow a}} < \frac{\beta(2-\alpha) N}{1-\alpha}.
    \label{eq:app_sanity_bound}
\end{equation}
\end{prf}
It is obvious that our cross-sanity check formulation (\cref{eq:app_sanity_bound}) is a relaxed version of the strict symmetry in \cref{eq:app_strict}. So far, we prove that under our cross-sanity check, the symmetry of $\tilde{g}^{b\rightarrow a}$ and $g^{a\rightarrow b}$ is bounded by $\alpha$ and $\beta$. When the strict symmetry is satisfied, our cross-sanity check is definitely satisfied. However, if our cross-sanity check is satisfied, it is not the other way around. We show in the experiments that our relaxed version of symmetry improves the overall results, quantitatively and qualitatively.

\subsection{Proof of Thm.~\ref{thm:minimizer}}
\label{sec:thm_minimizer}
\begin{theorem}[Existence of the unique minimizer for our relaxed optimization] 
Let $m$ and $f$ be two images defined on the same spatial domain $\Omega$, which is connected, closed, and bounded in $\mathbb{R}^n$ with a Lipschitz boundary $\partial \Omega$. Let $g: H\times H \rightarrow H^\star$ be a displacement mapping from the Hilbert space $H$ to its dual space $H^\star$. Then, there exists a unique minimizer $g^\star$ to the relaxed minimization problem.
\end{theorem}
\begin{prf}
In reality, a meaningful deformation field cannot be unbounded. We first restrict $g$ to be a closed subset of $L^2(H^\star)$:
\begin{equation}
\begin{aligned}[b]
    \mathcal{B}  \triangleq &\{g\in L^2(H^\star):\\
    &\normL{g}{2}^2\leq B, B \in \mathbb{R}_+ \text{ only depends on }\Omega\}.
\end{aligned}
\end{equation}
We then seek solutions $g^\star$ to the minimization problem in the space $H^1(\Omega)\cap \mathcal{B}$, and meanwhile satisfying our proposed checks. For short notations, we denote the minimization problem as 
\begin{equation}
    \begin{aligned}[b]
        &\qquad\qquad\qquad\min_{g\in H^1\cap \mathcal{B}} E(g), \\
        \qquad&\st \\
        &\mkern-24mu E(g)=-\Sim (f, m \circ (g^{m\rightarrow f}+id)) + \lambda_r\normtwotwo{\Reg(g^{m\rightarrow f})},
    \end{aligned}
\end{equation}
and $\lambda_r$ is a positive constant. For $g \in H^1\cap \mathcal{B}$, $E(g)$ is bounded below and there exists a minimizing sequence $\{g_k\}_{k=1}^\infty$ satisfying 
\begin{equation}
    \begin{aligned}[b]
        E(g_{k+1})\leq E(g_{k}) \leq \cdots \leq E(g_{1}) &\leq \lim_{k\rightarrow \infty}E(g_k)\\
        =\inf_{H^1\cap B}E(g).
    \end{aligned}
    \label{eq:bounded_seq}
\end{equation}
For identical inputs, we have $\normtwotwo{g}=0 < B\Rightarrow g \in H^1\cap\mathcal{B}$, thus, the self-sanity is checked. For different image pairs, we have 
\begin{equation}
    \begin{aligned}[b]
       \normtwotwo{g+\tilde{g}} &< \alpha(\normtwotwo{g} + \normtwotwo{\tilde{g}})+\beta N\leq 2\alpha B+\beta N.
    \end{aligned}
    \label{eq:cross_bound}
\end{equation}
By definition, $g$ and its reversed displacement $\tilde{g}$ need to follow the constraint $g(x)\tilde{g}(x)\leq 0, \forall x \in \Omega$, so we have
\begin{equation}
    \max(\normtwotwo{g+\tilde{g}}) < \normtwotwo{g} \leq B \Rightarrow 2\alpha B + \beta N < B.
\end{equation}
Here, we let $2\alpha B + \beta N < B$, since our cross-sanity check is considered a tighter bound than $B$. Thus, by choosing appropriate $\alpha$ and $\beta$, we can ensure that the cross-sanity is also checked, such that $g \in H^1\cap\mathcal{B}$. Due to the fact that $H^1$ is precompact in $L^2$ space, there exists a convergent subsequence where we still denote as $\{g_k\}_{k=1}^\infty$, and $g^\star \in H^1$, such that $g_k \rightarrow g^\star$, which is strongly in $L^2$ and \text{a.e.} in $\Omega$. Note that, either similarity functions (e.g. NCC) or distance functions (e.g. SSD) is naturally bounded in our image registration scenario, so that we can always have
\begin{equation}
    -\Sim (f, m \circ (g^{\star}+p)) \leq \lim_{k\rightarrow\infty}-\Sim (f, m \circ (g_k+p)).
    \label{eq:lim_sim}
\end{equation}
Besides, for the $H^1$ regularization, since $\{g_k\}_{k=1}^\infty$ is a bounded convergent sequence in $H^1\cap \mathcal{B}$, and $g_k \rightarrow g^\star$ \text{a.e.} in $\Omega$. By the dominant convergence theorem, we have
\begin{equation}
    \lim_{k\rightarrow \infty}\Reg(g_k)=\Reg(g^\star).
    \label{eq:lim_reg}
\end{equation}
Combining \cref{eq:lim_sim} with \cref{eq:lim_reg}, we obtain
\begin{equation}
    E(g^\star) \leq \lim_{k\rightarrow \infty}E(g_k)=\inf_{H^1\cap B}E(g).
\end{equation}
Thus, $g^\star$ is indeed a solution to the minimization problem. So far, we prove that there exists a minimizer $g^\star$ of the modified optimization problem. We can then prove that $g^\star$ is unique. Note that, we assume that the similarity operator is concave (e.g., NCC, negative NCC is convex) when it is not a distance operator (e.g., SSD is convex) on the transformation with a pair of $(m, f)$. I.e., $g^\star=g|m,f$, i.e., $g^\star$ learned by the model and conditioned on this specific $(m, f)$ pair to satisfy the proposed sanity checks. Therefore, since the data term (e.g. SSD) and the regularization ($H^1$ regularization) are convex, together with the
convex search space for $g$, the uniqueness of the minimizer $g^\star$ is proved.
\end{prf}

\subsection{Proof of Thm.~\ref{thm:loyalty}}
\label{sec:thm_loyalty}
\begin{theorem}[Loyalty of the sanity-checked minimizer]
Let $\rvg_\ast$ be the optimal minimizer to the bidirectional registration problem, defined in \cref{eq:san_optimal}, and $\rvg_{\rm sanity}$ as our sanity-checked minimizer, defined in \cref{eq:san_sanity}. The distance between these two minimizers can be upper bounded as
\begin{equation*}
\begin{aligned}[b]
    \Sim(\rvg_{\rm sanity})-\Sim(\rvg_{\ast})\leq \frac{\mathbf{\lambda}}{2}\normtwotwo{\rmA(\rvg_{\rm sanity}-\rvg_{\ast})}.
\end{aligned}
\end{equation*}
\end{theorem}
\begin{prf}
Since $\rvg_{\ast}$ is optimal, thus we have that $-\Sim(\rvg_{\ast})+\frac{\mathbf{\lambda}}{2}\normtwotwo{\rmA\rvg_\ast-\rvy}\leq-\Sim(\rvg_{\rm sanity})+\frac{\mathbf{\lambda}}{2}\normtwotwo{\rmA\rvg_{\rm sanity}-\rvy}$. Since $\normtwotwo{\rmA\rvg_\ast-\rvy}=0$, with elimination we have that $-\Sim(\rvg_{\ast})\leq-\Sim(\rvg_{\rm sanity})+\frac{\mathbf{\lambda}}{2}\normtwotwo{\rmA\rvg_{\rm sanity}-\rvy}$. By substituting $\rvy$ with $\rmA\rvg_\ast$ and combining like terms, we have~\cref{eq:san_distance}. Therefore, we prove the loyalty of the unique minimizer $\rvg_{\rm sanity}$ to the optimal minimizer $\rvg_\ast$, controlled by weight $\mathbf{\lambda}$.
\end{prf}

\subsection{Proof of CS Error Upper Bound}
\label{sec:prf_cs_bound}
Recall we want to prove that the CS error is upper bounded in the form of 
\begin{equation*}
    \begin{aligned}[b]
        \normtwotwo{\text{CS}(\rvg_{\rm sanity})} < 2(1-\alpha)\beta N.
    \end{aligned}
\end{equation*}
\begin{prf}
Recall CS error in \cref{eq:san_mini_bichecked_simple}, and by Thm.~\ref{thm:relaxation}, we have
\begin{equation}
\begin{aligned}[b]
    &\normtwotwo{\text{CS}(\rvg_{\rm sanity})}\\
    &=\normtwotwo{\rvg_{\rm sanity}+\tilde{\rvg}_{\rm sanity}} - \alpha(\normtwotwo{\rvg_{\rm sanity}} + \normtwotwo{\tilde{\rvg}_{\rm sanity}})-2\beta N \\
    &=(1-\alpha)(\normtwotwo{\rvg_{\rm sanity}} + \normtwotwo{\tilde{\rvg}_{\rm sanity}}\\
    &\quad\quad\quad\quad\quad\quad+\frac{2}{1-\alpha}{\rvg_{\rm sanity}}^\top\tilde{\rvg}_{\rm sanity})-2\beta N\\
    &\leq (1-\alpha)(\normtwotwo{\rvg_{\rm sanity}} + \normtwotwo{\tilde{\rvg}_{\rm sanity}} \\
    &\quad\quad\quad\quad\quad\quad+ 2{\rvg_{\rm sanity}}^\top\tilde{\rvg}_{\rm sanity})-2\beta N\\
    &= (1-\alpha)\normtwotwo{\rvg_{\rm sanity}+\tilde{\rvg}_{\rm sanity}}-2\beta N\\
    &< (1-\alpha)\frac{2\beta(2-\alpha)}{1-\alpha}N-2\beta N\\
    &=2(1-\alpha)\beta N, \quad \where 0<\alpha< 1 \quad \text{and} \quad \beta>0.
\end{aligned}
\end{equation}
The first inequality holds since ${\rvg_{\rm sanity}}^\top\tilde{\rvg}_{\rm sanity} \leq 0$ for valid inverse consistent displacements. The second inequality holds for two directions ($m{\rightarrow} f$ and $f{\rightarrow} m$) by Thm.~\ref{thm:relaxation} . Thus, the proof for CS error upper bound is completed.
\end{prf}

\begin{table}[!t]\footnotesize
    \centering
    \scalebox{0.88}{\begin{tabular}{llcccccr}
    \toprule
    $\alpha$ & $\beta$ & Dice & SDice & FV & \text{AJ}\texttimes{10}\textsuperscript{4} & \text{SSE}\texttimes{10}\textsuperscript{-1} &\multicolumn{1}{c}{CSE}\\
    \cmidrule(r){1-2}\cmidrule(r){3-4}\cmidrule(r){5-6}\cmidrule(r){7-8} 
    $^\ast$0.1 & 5 & 0.7263 & 0.8850 & 1.5220& 3.00 &5.156 & \multicolumn{1}{c}{10.72}\\
    \midrule
    $^\dag$0.1 & 5 &0.7223 & 0.8745& 1.6130& 3.30& 4.337&\multicolumn{1}{c}{11.05}\\
    \midrule
    0 & 0 &0.7178 & 0.9486& 1.4041& 2.74& 0.693&7.58\textrightarrow 1.81\textsubscript{76.12\%}\\
    \midrule
    \multirow{12}{*}{0.1} & 0.1 & 0.7199& 0.9533& 1.2460& 2.30&0.721 &6.39\textrightarrow 2.21\textsubscript{65.41\%}\\
    & 1 & 0.7223& 0.9654& 1.2070& 2.02&0.643 &6.81\textrightarrow 4.60\textsubscript{32.45\%}\\
    & 3 & 0.7233 & 0.9685& 1.0160& 1.51& 0.596&8.70\textrightarrow 5.19\textsubscript{40.34\%}\\
    & 5 & 0.7247 & 0.9588 & 0.8216 & 1.09 &0.723 & 10.72\textrightarrow 5.03\textsubscript{53.07\%}\\
    & 7 & 0.7226& 0.9689& 0.6632& 0.77& 0.642&12.53\textrightarrow 4.52\textsubscript{63.92\%}\\
    & 8 & 0.7232 & 0.9600& 0.6152& 0.68& 0.668 & 13.39\textrightarrow 4.27\textsubscript{68.11\%}\\
    & 9 & 0.7226 & 0.9699 & 0.5534 & 0.58 & 0.572 & 14.25\textrightarrow 4.03\textsubscript{71.72\%}\\
    & 11 & 0.7206 & 0.9665& 0.4472& 0.41& 0.564& 15.83\textrightarrow 3.53\textsubscript{77.70\%}\\
    & 12 & 0.7211 & 0.9668& 0.4219 & 0.38& 0.572 & 16.55\textrightarrow 3.43\textsubscript{79.27\%}\\
    & 13 & 0.7186& 0.9742& 0.4278& 0.39& 0.543& 17.22\textrightarrow 3.39\textsubscript{80.31\%}\\
    & 14 & 0.7189 & 0.9678& 0.4182 & 0.37& 0.543 & 17.80\textrightarrow 3.24\textsubscript{81.80\%}\\
    & 16 & 0.7179 & 0.9681 & 0.4345& 0.40& 0.561& 18.79\textrightarrow 2.89\textsubscript{84.62\%}\\
    & 20 & 0.7124& 0.9693 & 0.4668& 0.44&0.577 & 20.36\textrightarrow 2.82\textsubscript{86.15\%}\\
    \midrule
    0.01 & \multirow{6}{*}{5} & 0.7229& 0.9677& 0.9842& 1.46& 0.623&11.30\textrightarrow 5.43\textsubscript{51.94\%}\\
    0.10 & & 0.7247 & 0.9588 & 0.8216 & 1.09 &0.723 & 10.72\textrightarrow 5.03\textsubscript{53.07\%}\\
    0.15 & & 0.7230 & 0.9624 & 0.7891 & 0.99& 0.591  &10.42\textrightarrow 4.77\textsubscript{54.22\%}\\
    0.20 & & 0.7222 & 0.9656 & 0.7479 & 0.90& 0.586 & 10.15\textrightarrow 4.53\textsubscript{55.37\%}\\
    0.50 & & 0.7218 & 0.9570 & 0.7276 & 0.86 & 0.642&8.64\textrightarrow 3.47\textsubscript{59.83\%}\\
    1 & & 0.7165 & 0.9353 & 0.9622 & 1.45& 1.277& 7.41\textrightarrow 1.93\textsubscript{73.95\%}\\
    \bottomrule
    \end{tabular}}
    \caption{Ablation study for $\alpha, \beta$, where the loss weights for both sanity check losses are set to 0.001. $\ast$ denotes the checkpoint model used in this ablation (resulting in initial CSE errors on the left-side of arrows), and $\dag$ represents where we follow the standard protocol to finetune the model for the same epochs as the rest ablative experiments. In this case, $\alpha$ and $\beta$ are only used to calculate the CSE. Along with SSE, they are not part of the training loss.}
    \label{tab:abs_alpha}
\end{table}

\begin{figure}[!t]
    \centering
    \includegraphics[width=\linewidth]{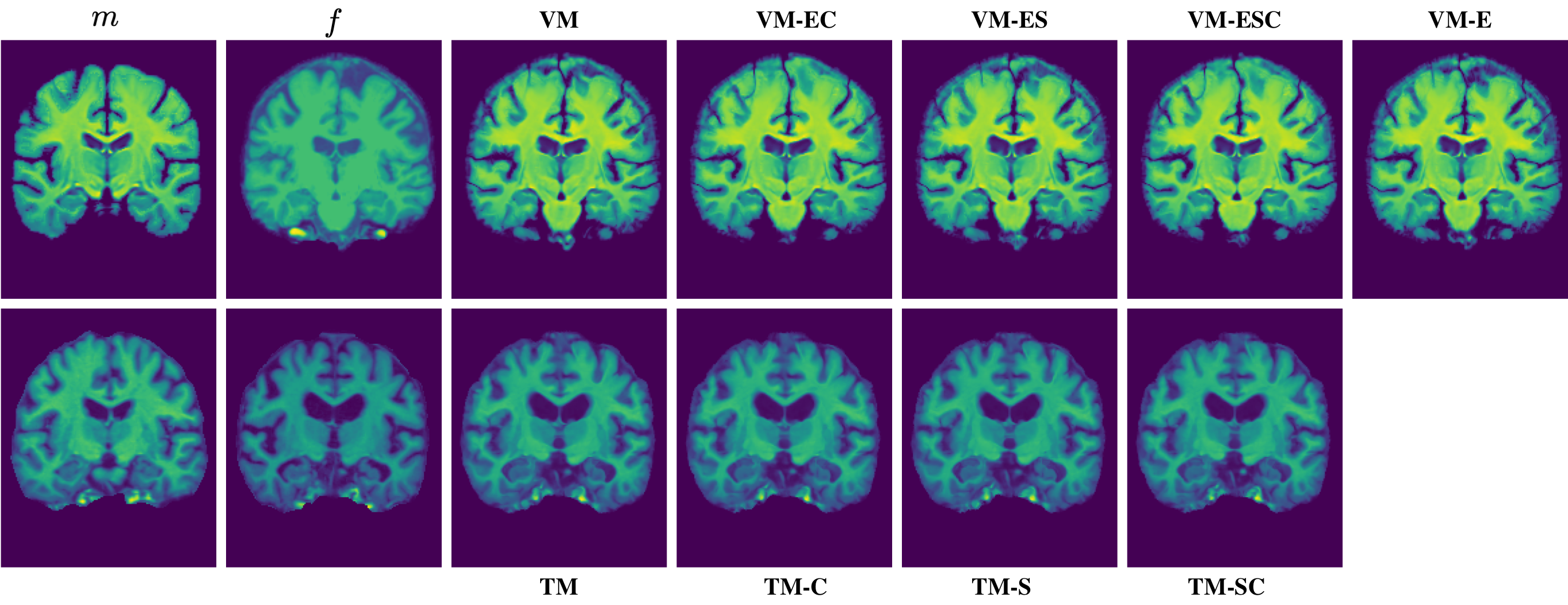}
    \caption{Qualitative comparisons of different model variants on: (\textit{Top}) IXI dataset, (\textit{Bottom}) OASIS validation dataset.}
    \label{fig:ablation}
\end{figure}
\subsection{Cross-sanity Check Numerical Study}
We present our numerical study for $\alpha$ and $\beta$ parameters in \cref{tab:abs_alpha}. This ablation study is conducted on the same subset of IXI training dataset described in the ablation study section but validate/test in the entire validation/test set. Our thought is that compared to the problem of enforcing strict inverse consistency over two different displacements, the relaxed version might be easier to solve, mathematically. Again, practically speaking, it is rather difficult to find one-to-one correspondence for every point in the moving-fixed image pair, which is the strict inverse consistency saying, relaxing or setting an error threshold can be effective from this perspective. Our error-bound formulation implicitly presents guidance for training such sanity-checked models.

\noindent\textbf{Settings of $\alpha$ and $\beta$.} A uniform estimate of $\alpha$ and $\beta$ is possible, however, such a bound is not sharp, and it will lead to over-estimation of $\lambda_c$ (the regularization parameter) for different applications. Hence, we prefer to derive the bounds on $\alpha$ and $\beta$ on particular sets of applications, where we can easily find such bounds from the formulations. As stated in \textbf{Thm. 1}, $\alpha$ and $\beta$ control the relaxation. We can also directly derive an upper bound from the check that constrains the ratio of two displacements. Since $g^{a\rightarrow b}\tilde{g}^{b\rightarrow a}{<}0$, then $\frac{g^{a\rightarrow b}}{\tilde{g}^{b\rightarrow a}}{+}\frac{\tilde{g}^{b\rightarrow a}}{g^{a\rightarrow b}}{<}\frac{2}{1-\alpha}$, $\beta$ is neglected for simplicity. These two bounds estimate ranges of $\alpha$ and $\beta$ for our relaxation. E.g., we use existing models (e.g., VM) to predict ten samples randomly, and set $\beta$ to 0.15\texttimes~maximum displacement; set $\alpha$ to 0.1 for models outputting absolute displacements (e.g., VM and TM), or $\alpha$ to 0.01 for models outputting relative displacements (e.g., DIRAC). Note that this only needs to be done once, while the previous experiment shows that the registrations are pretty robust among a range of $\alpha$ and $\beta$. Thus, it is safe to choose $\alpha$ and $\beta$ within the range.
\begin{table}[t]\footnotesize
    \centering
    \scalebox{0.8}{\begin{tabular}{llcccccccc}
    \toprule
    \scalebox{1.2}{$\frac{\text{Train}}{\text{Test}}$} & \scalebox{1.0}{Model} & \scalebox{1.0}{Dice} & \scalebox{1.0}{SDice} & \scalebox{1.0}{HD95} & \scalebox{1.0}{SDlogJ} & \scalebox{1.0}{FV} & 
    \scalebox{0.8}{\text{AJ}\texttimes{10}\textsuperscript{4}} & 
    \scalebox{0.8}{\text{SSE}\texttimes{10}\textsuperscript{-1}} & \scalebox{1.0}{CSE}\\
    \cmidrule(r){1-1}\cmidrule(r){2-2}\cmidrule(r){3-6}\cmidrule(r){7-8}\cmidrule(r){9-10}
    \scalebox{1.2}{\multirow{4}{*}{$\frac{\text{IXI}}{\text{OASIS}}$}} & \scalebox{0.95}{VM} & 0.714&0.848 & 3.039& 0.118& 0.831&1.52 &6.24&7.49\\
    & \scalebox{0.95}{\textit{VM-ESC}} & 0.759 & 1.000 & 2.563& 0.080&0.173 &0.16& 0.00& 2.86\\
    \cmidrule(l){2-10}
    & \scalebox{0.95}{TMBS} & 0.775&0.901 &2.201 &0.072 &0.00 &0.00 & 24.4&15.16\\
    & \scalebox{0.95}{\textit{TMBS-ES}C} & 0.783& 1.000 & 2.144&0.061 & 0.00 & 0.00 & 0.00 &3.73\\
    \midrule
    \scalebox{1.2}{\multirow{2}{*}{$\frac{\text{OASIS}}{\text{IXI}}$}} & \scalebox{0.95}{TM} & 0.705& 0.918& 4.083& 0.132&1.390 & 2.67 & 3.18 & 13.78\\
    & \scalebox{0.95}{\textit{TM-SC}} & 0.718& 1.000& 3.903&0.097 &0.674 &0.84 & 0.01 & 5.58\\
    \bottomrule
    \end{tabular}}
    \caption{Generalization ability study on sanity-checked registers to study cross-dataset registration performance. Here, we set $\alpha{=}0.1$, and $\beta{=}12$ to calculate cross-sanity errors in both settings.}
    \label{tab:generalization}
\end{table}
\begin{figure}[!t]
    \centering
    \includegraphics[width=\linewidth]{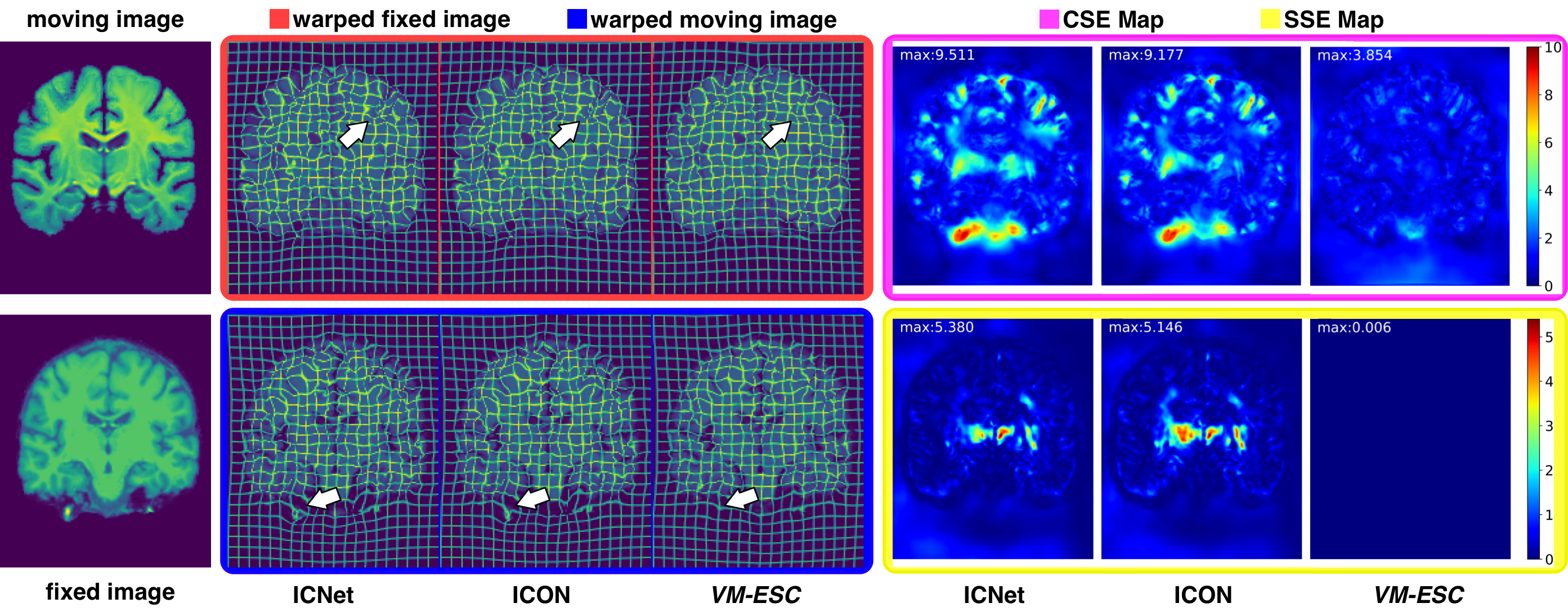}
    \caption{Qualitative comparisons between ICNet~\cite{zhang2018inverse}, ICON~\cite{Greer_2021_ICCV} and our method on IXI dataset.}
    \label{fig:inv_comp}
\end{figure}

\subsection{Ablative Qualitative Comparisons}
Qualitative comparisons between our ablative model variants are shown in \cref{fig:ablation}.

\subsection{Sanity-awareness Preservation Study on Cross-Dataset Scenario} We test whether sanity awareness is preserved in cross-dataset scenarios. We train our sanity-checked register in one dataset and then test it on different dataset so no overlapping between training and testing datasets. The results are shown in \cref{tab:generalization}. Compared to methods without sanity checks, our sanity-checked models improve in every metric, certifying that our sanity checks do not harm the model training. Besides, the sanity-checked registers still preserve good sanity for preventing corresponding errors.

\subsection{Experimental Results Statistical Significance}

We specifically study whether our results are statistically significant, compared to the other strong baselines, e.g., ICON~\cite{Greer_2021_ICCV} and DIRAC~\cite{mok2022unsupervised}. We calculate \texttt{p value} using \texttt{scipy} package. Compared to ICON, our \textit{VM-ESC} (\texttt{p value: 0.0174}) and \textit{TMBS-ESC} (\texttt{p value: 0.0280}), while for DIRAC, our \textit{DIRAC-SC} (\texttt{p value: 0.0204}), considering all metrics shown in the corresponding tables. All the \texttt{p values} are $<0.05$, indicating that our results are statistically significant.

\subsection{Error Map Comparisons between Inverse Consistent Methods}
Qualitative comparisons between inverse consistent methods on IXI dataset are shown in \cref{fig:inv_comp}.

\begin{table}[!t]\small
    \centering
    \scalebox{0.7}{\begin{tabular}{|c|c|c|c|c|c|}
    \hline
    Method & TRE\textdownarrow & STRE\textdownarrow & ROB\textuparrow& FV\textdownarrow & AJ\texttimes{10}\textsuperscript{2}\textdownarrow\\
    \hline
    DIRAC & 2.760$\pm$0.247&0.274$\pm$0.027& 0.776$\pm$0.055& 0.025$\pm$0.009&4.242$\pm$2.954 \\
    \hline
    \textit{DIRAC-C} & 2.721$\pm$0.262&0.268$\pm$0.039& 0.791$\pm$0.044& 0.022$\pm$0.008&3.012$\pm$1.442 \\
    \textit{DIRAC-SC} &2.719$\pm$0.259&0.218$\pm$0.046 & 0.795$\pm$0.034& 0.022$\pm$0.005&3.001$\pm$1.314 \\
    \hline
    \end{tabular}}
    \caption{Performance of replacing the inverse consistent error.}
    \label{tab:app_dirac}
\end{table}
\subsection{Performance of Replacing DIRAC's Inverse Error} 
We denote it as \textit{DIRAC-C}, and report in \cref{tab:app_dirac}.

\subsection{Role of Image Similarity Loss} 
The image similarity loss still plays a very important role during training. The reason is that $\mathcal{L}_{\rm self}$ and $\mathcal{L}_{\rm cross}$ are defined on displacements, to calculate such losses, we need to ensure that those displacements are meaningful, which is guaranteed via $\mathcal{L}_{\rm sim}$. Compared to the value of NCC ($<$1), the cross-sanity error is relatively large (Tab.~\ref{tab:IXI}), and using small $\lambda_c$ will not interfere with the optimizations.

\end{document}